\documentclass[twoside]{article}
\usepackage[accepted]{aistats2019}

\usepackage[round]{natbib}

\usepackage[utf8]{inputenc} % allow utf-8 input
\usepackage[T1]{fontenc}    % use 8-bit T1 fonts
\usepackage{lmodern}
\usepackage{hyperref}       % hyperlinks
\usepackage{url}            % simple URL typesetting
\usepackage{booktabs}       % professional-quality tables
\usepackage{amsfonts}       % blackboard math symbols
\usepackage{nicefrac}       % compact symbols for 1/2, etc.
\usepackage{microtype}      % microtypography

\usepackage{setspace}
\usepackage{geometry}
\usepackage{graphicx}
\usepackage{xcolor}
\usepackage{titlesec}
\usepackage[page]{appendix}
\usepackage{enumerate}
\usepackage{stmaryrd}
\usepackage{bookmark}
\usepackage{lscape}

\usepackage{lscape}
\usepackage{booktabs}
\usepackage{rotating}
\usepackage{multirow}
\usepackage{longtable}
\usepackage{caption}
\usepackage{subcaption}
\usepackage{float}
\usepackage{tabularx}
\usepackage{ragged2e}
\newcolumntype{Y}{>{\RaggedRight\arraybackslash}X}
\usepackage{pdflscape}
\usepackage{afterpage}

\usepackage{amsmath}
\usepackage{amssymb}
\usepackage{amsthm}
\usepackage{mathtools}
\usepackage{dsfont}

\usepackage[ruled,vlined]{algorithm2e}

\SetCommentSty{algcommstyle}

\usepackage{tikz}
\usetikzlibrary{bayesnet}

\usepackage{textcomp}
\usepackage{sourcecodepro}
\usepackage{listings}
\definecolor{commentgrey}{gray}{0.45}
\definecolor{backgray}{gray}{0.96}
\DeclareMathOperator*{\argmax}{arg\,max}
\DeclareMathOperator*{\argmin}{arg\,min}
\DeclareMathOperator*{\tr}{tr}

\DeclarePairedDelimiter{\norm}{\lVert}{\rVert}

\newcommand*{\cond}{\;\ifnum\currentgrouptype=16 \middle\fi|\;}

\newcommand*{\m}[1]{\textbf{#1}}

\newcommand*{\ttilde}{{\raise.17ex\hbox{$\scriptstyle\sim$}}}

\makeatletter
\newsavebox{\mybox}\newsavebox{\mysim}
\newcommand*{\distas}[1]{%
  \savebox{\mybox}{\hbox{\kern3pt$\scriptstyle#1$\kern3pt}}%
  \savebox{\mysim}{\hbox{$\sim$}}%
  \mathbin{\overset{#1}{\kern\z@\resizebox{\wd\mybox}{\ht\mysim}{$\sim$}}}%
}
\makeatother

\makeatletter
\def\moverlay{\mathpalette\mov@rlay}
\def\mov@rlay#1#2{\leavevmode\vtop{%
   \baselineskip\z@skip \lineskiplimit-\maxdimen
   \ialign{\hfil$\m@th#1##$\hfil\cr#2\crcr}}}
\newcommand*{\charfusion}[3][\mathord]{
  #1{\ifx#1\mathop\vphantom{#2}\fi\mathpalette\mov@rlay{#2\cr#3}}
  \ifx#1\mathop\expandafter\displaylimits\fi}
\makeatother

\newcommand*{\mt}[1]{\text{\normalfont #1}}
\newcommand{\silent}[1]{}
\newcommand{\ignore}[1]{}{}

\newtheorem{theorem*}{Theorem}

\newtheorem{corollary*}{Corollary}
\newtheorem{proposition}{Proposition}[section]
\newtheorem{proposition*}{Proposition}
\newtheorem{lemma}{Lemma}[section]
\newtheorem{lemma*}{Lemma}

\theoremstyle{definition}

\newtheorem{definition*}{Definition}

\newtheoremstyle{algodesc}{}{}{}{}{\bfseries}{.}{ }{}%
\theoremstyle{algodesc}

\begin{document}
% \title{Minimum Volume Topic Modeling}
% \maketitle
\twocolumn[

\aistatstitle{Minimum Volume Topic Modeling}

\aistatsauthor{Byoungwook Jang \And Alfred Hero}% Author 1 \And Author 2 \And  Author 3 }
% \vspace{15pt}
\aistatsaddress{
  {\small Department of Statistics} \\ {\small University of Michigan}\\{\small Ann Arbor, MI, 48109}\\{\small bwjang@umich.edu}
\And
  {\small Department of EECS} \\ {\small University of Michigan}\\{\small Ann Arbor, MI, 48109}\\{\small hero@umich.edu}} ]% Institution 1 \And  Institution 2 \And Institution 3 } ]

\begin{abstract}
We propose a new topic modeling procedure that takes advantage of the fact that the Latent Dirichlet Allocation (LDA) log likelihood function is asymptotically equivalent to the logarithm of the volume of the topic simplex. This allows topic modeling to be reformulated as finding the probability simplex that minimizes its volume and encloses the documents that are represented as distributions over words. A convex relaxation of the minimum volume topic model optimization is proposed, and it is shown that the relaxed problem has the same global minimum as the original problem under the separability assumption and the sufficiently scattered assumption introduced by \cite{arora2013practical} and \cite{huang2016anchor}. A locally convergent alternating direction method of multipliers (ADMM) approach is introduced for solving the relaxed minimum volume problem. Numerical experiments illustrate the benefits of our approach in terms of computation time and topic recovery performance.
\end{abstract}

\section{Introduction}
Since the introduction by \citet{blei_et_al_03} and \citet{pritchard2000inference}, the Latent Dirichlet Allocation (LDA) model has remained an important tool to explore and organize large corpora of texts and images. The goal of topic modeling can be summarized as finding a set of topics that summarizes the observed corpora, where each document is a combination of topics lying on the topic simplex.

There are many extensions of LDA, including a nonparametric extension based on the Dirichlet process called Hierarchical Dirichlet Process \citep{teh2005sharing}, a correlated topic extension based on the logistic normal prior on the topic proportions \citep{blei_lafferty_06}, and a time-varying topic modeling extension \citep{blei2006dynamic}. There are two main approaches for estimation of the parameters of probabilistic topic models: the variational approximation popularized by \citet{blei_et_al_03} and the sampling based approach studied by \citet{pritchard2000inference}. These inference algorithms either approximate or sample from the posterior distributions of the latent variable representing the topic labels. Therefore, the estimates do not necessarily have a meaningful geometric interpretation in terms of the topic simplex - complicating assessment of goodness of fit to the model. In order to address this problem, \citet{yurochkin_long_16} introduced Geometric Dirichlet Mean (GDM), a novel geometric approach to topic modeling. It is based on a geometric loss function that is surrogate to the LDA's likelihood and builds upon a weighted k-means clustering algorithm, introducing a bias correction. It avoids excessive redundancy of the latent topic label variables and thus improves computation speed and learning accuracy. This geometric viewpoint was extended to a nonparametric setting \citep{yurochkin2017conic}.

LDA-type models also arise in the hyperspectral unmixing problem. Similar to the documents in topic modeling, hyperspectral image pixels are assumed to be mixtures of a few spectral signatures, called endmembers (equivalent to topics). Unmixing procedures aim to identify the number of endmembers, their spectral signatures, and their abundances at each pixel (equivalent to topic proportions). One difference between topic modeling and unmixing is that hyperspectral spectra are not normalized. Nonetheless, algorithms for hyperspectral unmixing are similar to topic model algorithms, and similar models have been applied to both problems. Geometric approaches in the hyperspectral unmixing literature take advantage of the fact that linearly mixed vectors also lie in a simplex set or in a positive cone. One of the early geometric approaches to unmixing was introduced in \citet{nascimento2005vertex} and \citet{bioucas2009variable}, which aim to first identify the K-dimensional subspace of the data and then estimate the endmembers that minimize the volume of the simplex spanned by these endmembers. \citet{bioucas2009variable} estimates the endmembers through minimizing the log determinant of the endmember matrix, as the log-determinant is proportional to the volume of the simplex defined by the endmembers. This idea of minimizing the simplex volume motivated the algorithm proposed in this paper for topic modeling. In \citet{bioucas2009variable}, however, the authors experience an optimization issue as their formulation is highly non-convex. It was found that the local minima of the objective in \citet{bioucas2009variable} may be unstable.

The topic modeling problem also has similarities to matrix factorization. In particular, nonnegative matrix factorization, while it does not enforce a sum-to-one constraint, is directly applicable to topic modeling \citep{deerwester1990indexing,xu2003document,anandkumar2012spectral,arora2013practical,fu2018anchor}.
Recover KL, recently introduced by \citet{arora2013practical}, provides a fast algorithm that identifies the model under a separability assumption, which is the assumption that the sample set includes the vertices of the true topic model (pure endmembers). As the separabiiltiy assumption is often not satisfied in practice, \citet{fu2018anchor} introduced a weaker assumption called the sufficiently scattered assumption. We provide a theoretical justification of our geometric minimum value method under this weaker assumption.

\subsection{Contribution}
We propse a new geometric inference method for LDA that is formulated as minimizing the volume of the topic simplex. The estimator is shown to be identifiable under the separability assumption and the sufficiently scattered assumption. Compared to \citet{bioucas2009variable}, our geometric objective involves $\log \det \beta \beta^T$ instead of $\log |\det \beta|$, making our objective function convex. At the same time, the $\log \det \beta \beta^T$ term remains proportional to the volume enclosed by the topic matrix $\beta$ and simplifies the optimization. In particular, we propose a convex relaxation of the minimization problem whose global minimization is equivalent to the original problem. This relaxed objective function is minimized using an iterative augmented Lagrangian approach, implemented using the alternating direction method of multipliers (ADMM), that is shown to be locally convergent.

\subsection{Notation}
We use the following notations. We are given a corpus $W\in \mathbb{D}^{M\times V}$ with $M$ documents, $K$ topics, vocabulary size $V$ and $N_m$ words in document $m$ for $m=1, \cdots, M$. Let $\mathbb{D}^{n \times p}$ be the space of $n\times p$ row-stochastic matrices. Then, our goal is to decompose $W$ as $W = \theta \beta$, where $\theta \in \mathbb{D}^{M\times K}$ is the matrix of topic proportions, and $\beta \in \mathbb{D}^{K \times V}$ is the topic-term matrix. Finally, $\Delta^{d}$ represents the $d$-dimensional simplex. It is assumed that the documents in the corpus obey the following generative LDA model.
\vspace{-10pt}
\begin{enumerate}
    \item For each topic $\beta_i$ for $i=1, \cdots, K$
    \vspace{-5pt}
    \begin{enumerate}
        \item Draw a topic distribution $\beta_i$
    \end{enumerate}
    \vspace{-5pt}
    \item For $j$-th document $w^{(j)} \in \mathbb{R}^V$ in the corpus $W$ for $j=1, \cdots, M$
    \vspace{-5pt}
    \begin{enumerate}
        \item Choose the topic proportion $\theta_w \sim Dir(\alpha)$
        \item For each word $\delta_n$ in the document $w^{(j)}$
        \begin{enumerate}
            \item Choose a topic $z_n \sim Mult(\theta)$
            \item Choose a word $\delta_n\sim \beta_{z_n}$
        \end{enumerate}
    \end{enumerate}

\end{enumerate}
%%%%%%%%%%%%%%%%%%%%%%%%%%%%%%%%%%%%%%%%%%%%%%%
%%%% Proposed Method
%%%%%%%%%%%%%%%%%%%%%%%%%%%%%%%%%%%%%%%%%%%%%%%
\section{Proposed Approach}
We assume that the number $K$ of topics is known in advance and is much smaller than size of the vocabulary, i.e. $K \ll V$. Furthermore, since LDA models the document as being inside the topic simplex, it is advantageous to represent the documents in a $K$-dimensional subspace basis.

Let $E_K = [e_1, \cdots, e_K]$ be a matrix of dimension $V \times K$ with $K$ orthogonal directions spanning the document subspace. Specifically, we define $E_K$ as the set of $K$ eigenvectors of the sample covariance matrix of the documents $w^{(i)}$, $i=1, \ldots, M$.

Most of the paper focuses on working with $\widetilde{w}^{(i)} = w^{(i)} E_K  \in \mathbb{R}^K$, which corresponds to the coordinates of $w^{(i)}$ in $\mt{colspan}(E_K)$. Note that we can recover the projected documents in the original $V$-dimensional space by
\begin{align*}
    \widehat{w}^{(i)} & = \overline{w} + (w^{(i)} - \overline{w}) E_K E_K^T \\
    & = \overline{w} + (\widetilde{w}^{(i)} - \overline{w} E_K) E_K^T \in \mathbb{R}^V
    % = \overline{w}(I_V -E_K E_K^T) + \tilde{w}^{(i)} E_K^T \in \mathbb{R}^V
\end{align*}
where $\overline{w}$ is the sample average of the observed documents. Therefore,
\begin{align*}
    {W} =  \theta {\beta} \Rightarrow (W - \theta \beta)E_K = 0
\end{align*}
where $\Theta$ belongs to the simplex $\Delta^K$. This $K$-dimensional probability simplex is defined by the topic distributions, which are the rows of $\beta E_K \in \mathbb{R}^{K\times K}$. For the rest of the paper, given $\omega \in \mathbb{R}^V$, we denote $\widetilde{\omega}$ as the corresponding coordinates in the projected subspace and $\widehat{\omega}$ as the projected vector in the original $V$-dimensional space.

\subsection{Topic Estimation}
Let $\gamma = (\beta E_K)^{-1}$. Then, it follows that $\theta = (WE_K)\gamma$. We know that $\beta E_K$ is invertible as we assume that there are $K$ distinct topics, and the rank of the topic matrix $\beta$ is $K$. Then, as noted in \citet{nascimento2012hyperspectral}, the likelihood w.r.t. $\Theta$ can be written as
\begin{equation}
\label{eq:topic_likelihood}
    \begin{aligned}
        & l(\theta, \beta| W)  = \sum_{i=1}^M p(w^{(i)} | \beta, \alpha) \\
        & \quad = \sum_{i=1}^M \log\left (p( \theta^{(i)} = (w^{(i)} E_K) \gamma |\beta, \alpha) \cdot |\det(\gamma)|\right ) \\
        & \quad = \sum_{i=1}^M \log\left (p( \theta^{(i)} = (w^{(i)} E_K) \gamma|\beta, \alpha) \right ) \\
        & \qquad \qquad + M \log |\det(\gamma)| \\
        % & \quad = \sum_{i=1}^N \log\left (p( \theta^{(i)} = ( w^{(i)} E_K) \gamma|\theta, \beta, \alpha) \right ) \\
        % & \qquad \qquad- N \log |\det(\widetilde{\beta})| \\
    \end{aligned}
\end{equation}

This formulation gives a nice geometric interpretation.

\textbf{Geometric Interpretation of log likelihood:} As we increase the number of documents $M \rightarrow \infty$, the dominant term is $\log |\det(\beta)|$. That is,
\begin{equation}
\label{eq:geom_interpretation}
    \begin{aligned}
      & \lim_{N\rightarrow \infty} \argmax_{\beta} l(\theta, \beta|W) \vspace{-10pt}\\
      & \quad = \lim_{M\rightarrow \infty} \argmax_{\widetilde{\beta}} \sum_{i=1}^M \log p( W_i | \theta, \beta) \\
      & \quad \approx \argmin_{\beta} \log |\det (\beta E_K)|\\
      & \quad = \argmin_\gamma - \log |\det \gamma|
    \end{aligned}
\end{equation}
% where $B = \{ \widetilde{\beta} \in \mathbb{R}^{K\times K}: \widetilde{\beta} \textbf{1} = \textbf{1}, \beta > 0\}$, i.e. a set of $K\times K$ row stochastic matrices.
Note that $\log |\det (\beta E_K)|$ is proportional to the volume enclosed by the row vectors of $\beta E_K$, i.e. the topic simplex in the projected subspace. In other words, the estimated topic matrix $\beta$ that minimizes its intrinsic volume is asymptotically equivalent to the asymptotic form of the log-likelihood \eqref{eq:topic_likelihood}. This is the main motivation for our proposal to minimize the volume of topic simplex.

\section{Minimum Volume Topic Modeling}
\label{sec:topic_admm}
In the remote sensing literature, \citet{nascimento2012hyperspectral} proposed to work with the likelihood \eqref{eq:topic_likelihood} by modeling $\theta$ as a Dirichlet mixture. However, their endmembers are spectra and do not necessarily satisfy the sum-to-one constraints on the endmember matrix; constraints which are fundamental to topic modeling. These additional constraints on the endmember complicate the minimization of \eqref{eq:topic_likelihood}. The first difficulty arises from the $\log| \det \beta|$ term, as $\beta$ is not a symmetric matrix, which makes the log likelihood $\eqref{eq:topic_likelihood}$ non-convex. Due to this non-convexity issue, \citet{nascimento2012hyperspectral} propose using a second-order approximation to the $\log \det \beta$ term. Yet, no rigorous justification has been provided to their approach. In contrast, we propose using $\log \det \beta \beta^T$ instead of $\log |\det \beta|$, prove identifiability under the sufficiently scattered assumption, and derive an ADMM update.

As we are optimizing $(\beta E_K)^{-1}$ directly, we use the notation $\gamma = (\beta E_K)^{-1}$ in the sequel. We can then rewrite the objective \eqref{eq:geom_interpretation} as follows
\begin{equation}
\label{eq:geom_topic_obj}
    \begin{aligned}
        \widehat{\gamma} &= \argmin_{\gamma\in \mathbb{R}^{K\times K}} -\log|\det(\gamma \gamma^T)| \\
        & \mt{s.t.}  \quad \theta > 0 \quad \theta \textbf{1} = \textbf{1} \quad \theta = (WE_K)\gamma \\
        & \quad \beta > 0 \quad \beta \textbf{1} = \textbf{1}
    \end{aligned}
\end{equation}
where $\beta = \overline{w} + (\gamma^{-1} - \overline{w}E_K)E_K^T$. The first set of constraints corresponds to the sum-to-one and nonnegative constraint on the topic proportions $\theta = (W E_K) \gamma$, and the second constraint imposes the same conditions on $\beta$. Thus, the problem \eqref{eq:geom_topic_obj} provides an exact solution to the asymptotic estimation of \eqref{eq:topic_likelihood}. However, this is not a convenient formulation of the optimization problem, as it involves the constraint on the inverse of $\gamma$. Note that as we assume $\beta \in \mathbb{R}^V$ intrinsically lives in a $K$-dimensional subspace, there is a one-to-one mapping between $\beta$ and $\gamma = (\beta E_K)^{-1}$. Throughout this paper, we will make use of this relationship between $\beta$ and $\gamma$.
% However, as there is a constraint on the inverse of the optimization variable, this problem is infeasible.
Here, working with a geometric interpretation of the second set of constraints we propose a relaxed version of \eqref{eq:geom_topic_obj}.

\textbf{Sum-to-one constraint on $\beta$:} Combined with the non-negativity constraint, the sum-to-one constraint $\beta \textbf{1} = \textbf{1}$ forces the rows of $\beta$ to lie in the $K$-dimensional topic simplex within the word simplex. To be specific, $\beta \textbf{1} = 1$ narrows our search space to be in an affine subspace, which is accomplished with a projection of the documents onto this $K$-dimensional affine subspace. This projection takes care of the sum-to-one constraint in the objective \eqref{eq:geom_topic_obj}.

\textbf{Non-negativity constraint on $\beta$:} We propose relaxing the non-negativity constraint to the following
\begin{equation*}
    \sigma_{min}(\gamma) \geq R^{-1}
\end{equation*}
where $\sigma_{min}(\gamma)$ is the minimum singular value of $\gamma$. As illustrated in Figure 1, this is interpreted as replacing the non-negativity constraint on the elements of the matrix $\beta$ with a radius $R$ ball constraint on the rows of the matrix $\beta$. As noted before, there is a mapping between $\gamma$ and $\beta$ through $\beta = \overline{w} + (\gamma^{-1} - \overline{w}E_K)E_K^T$. Thus, if $\gamma^t$ is the current iterate of an iterative optimization algorithm, to be specified below, then we can represent the corresponding $i$-th topic vector in the projected space as $b_i^t = (\gamma^t)^{-1}[i,:] = (\beta^t E_K)[i,:]$. It follows that
\begin{equation}
\label{eq:gamma_constraint}
\begin{aligned}
    \norm{b_i^t}^2 & = \frac{ \tr((b_i^t)^T b_i^t) \lambda_{min}(\gamma \gamma^T)}{\lambda_{min}(\gamma \gamma^T)}
     \leq \frac{\tr((b_i^t)^T b_i^t \gamma \gamma^T)}{\lambda_{min}(\gamma \gamma^T)} \\
    & = \frac{\tr(b_i^t \gamma (b_i^t \gamma)^T)}{\lambda_{min}(\gamma \gamma^T)}
     = \frac{\tr(e_i e_i^T)}{\sigma_{min}(\gamma)^2} \leq R^2
\end{aligned}
\end{equation}
Then, imposing $\sigma_{min}(\gamma ) \geq R^{-1}$ results in $\norm{b_i}^2 \leq R$. The first inequality in \eqref{eq:gamma_constraint} comes from the fact that $\lambda_{min} (A) \tr(B) \leq \tr(A B) \leq \lambda_{max}(A) \tr(B)$ for positive semidefinite matrices $A$ and $B$. The second equality in \eqref{eq:gamma_constraint} comes from the definition that $b_i$ is the $i$-th row of $(\gamma^t)^{-1}$

\begin{figure}[h]
    \centering
    \includegraphics[width=0.5\linewidth]{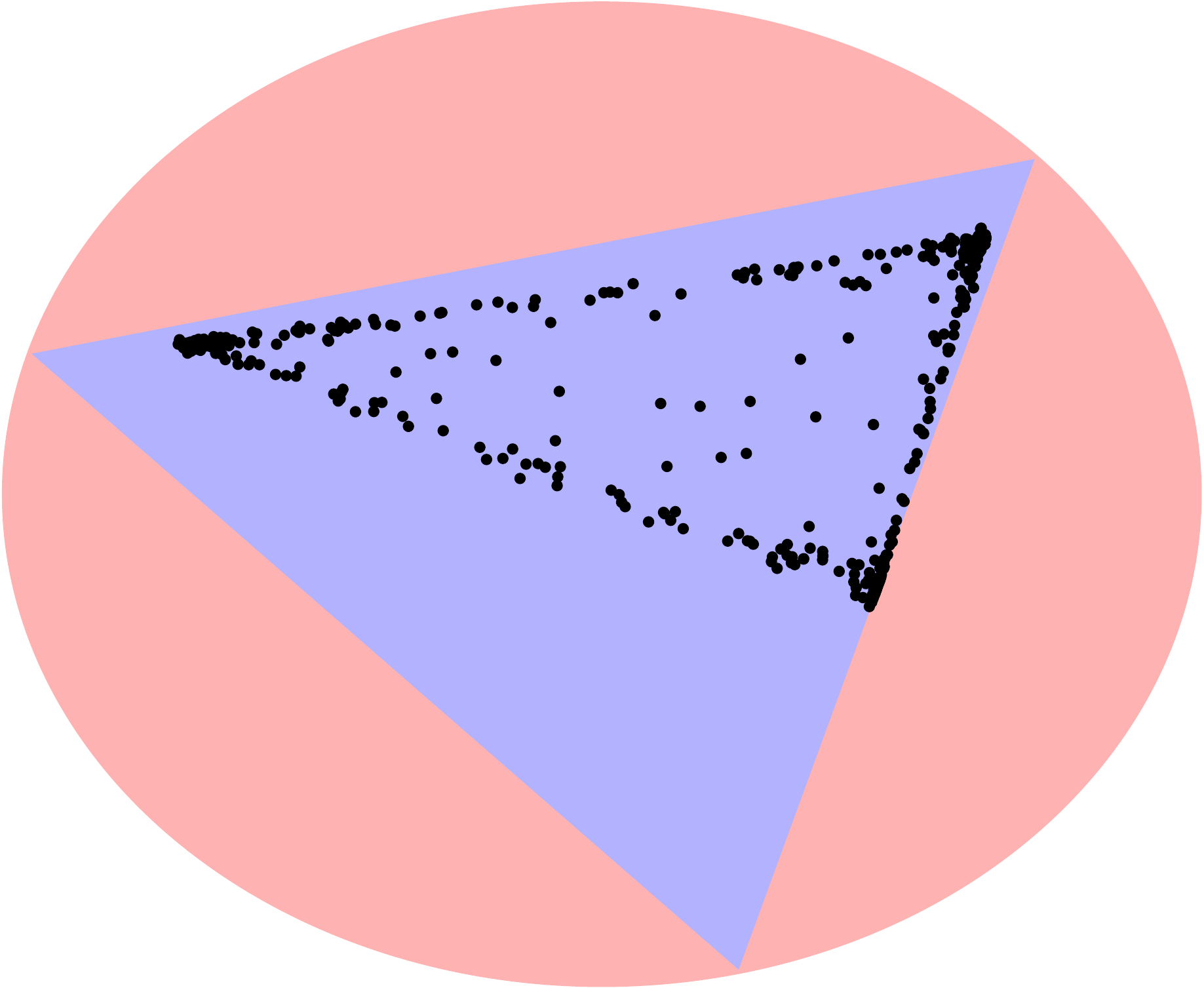}
    \caption{Visualization of the difference between the feasible space of \eqref{eq:geom_topic_obj} (blue triangle) and that of \eqref{eq:geom_topic_relaxed} (pink circle).}
    \label{fig:relaxed_space}
\end{figure}

With this spectral relaxation of the non-negativity constraint, the relaxed version of the problem \eqref{eq:geom_topic_obj} becomes
\begin{equation}
\label{eq:geom_topic_relaxed}
    \begin{aligned}
        \widehat{\gamma} &= \argmin_\gamma -\log|\det(\gamma \gamma^T)|\\
        & \mt{s.t.}  \quad \theta > 0 \quad \theta \textbf{1} = \textbf{1} \quad \theta = (WE_K)\gamma \\
        & \qquad \sigma_{min}(\gamma) \geq R^{-1}
        % \gamma \textbf{1}_K = (\widetilde{W}^T \widetilde{W})^{-1} \widetilde{W}^T \textbf{1}_V \quad \lambda_{min}(\gamma \gamma^T) \geq \frac{1}{\widetilde{R}^2} \\
    \end{aligned}
\end{equation}
Intuitively, as shown in Figure~\ref{fig:relaxed_space}, the optimization problem \eqref{eq:geom_topic_obj} and \eqref{eq:geom_topic_relaxed} are equivalent to each other except that the ball relaxation has expanded hthe solution space beyond the feasible space.

The blue triangle represents the set of feasible points for problem \eqref{eq:geom_topic_obj}, and the red circle corresponds to the solution space in \eqref{eq:geom_topic_relaxed}.

\subsection{Identifiability}

Here we establish identifiability of the model obtained by solving problem \eqref{eq:geom_topic_relaxed}. Identifiability gained interests in the topic modeling literature (\citet{arora2013practical} and \citet{fu2018anchor}). We show the identifiability under the sufficiently scattered condition. We first state the following lemma.

% \begin{equation}
%     B = \{\beta \in \mathbb{D}^{K\times V}: \exists \theta \in \mathbb{D}^{M \times K} \mt{ s.t. } W = \theta \beta\}
% \end{equation}
\begin{lemma}
\label{lemma:sol_space}
    Let $\widehat \gamma$ be a solution to the problem \eqref{eq:geom_topic_relaxed}. If $rank(W)=K$, we have that $\widehat{\gamma} \in \Gamma$, where
    % \begin{equation*}
    %     \beta = \overline{w} + (\gamma^{-1} - \overline{w}E_K)E_K^T
    % \end{equation*}
    \begin{equation*}
        \begin{aligned}
            & \Gamma = \{\gamma \in \mathbb{R}^{K\times K}:\beta = \overline{w} + (\gamma^{-1} - \overline{w}E_K)E_K^T \in \mathbb{D}^{K\times V} \\
            &  \quad \qquad \qquad \qquad \mt{ and } \exists \theta \in \mathbb{D}^{M \times K} \mt{ s.t. } \theta = (WE_K)\gamma\}
        \end{aligned}\textbf{}
    \end{equation*}
\end{lemma}

\begin{figure}[h]
    \begin{subfigure}[t]{0.45\linewidth}
        \centering
        \includegraphics[width=0.9\linewidth]{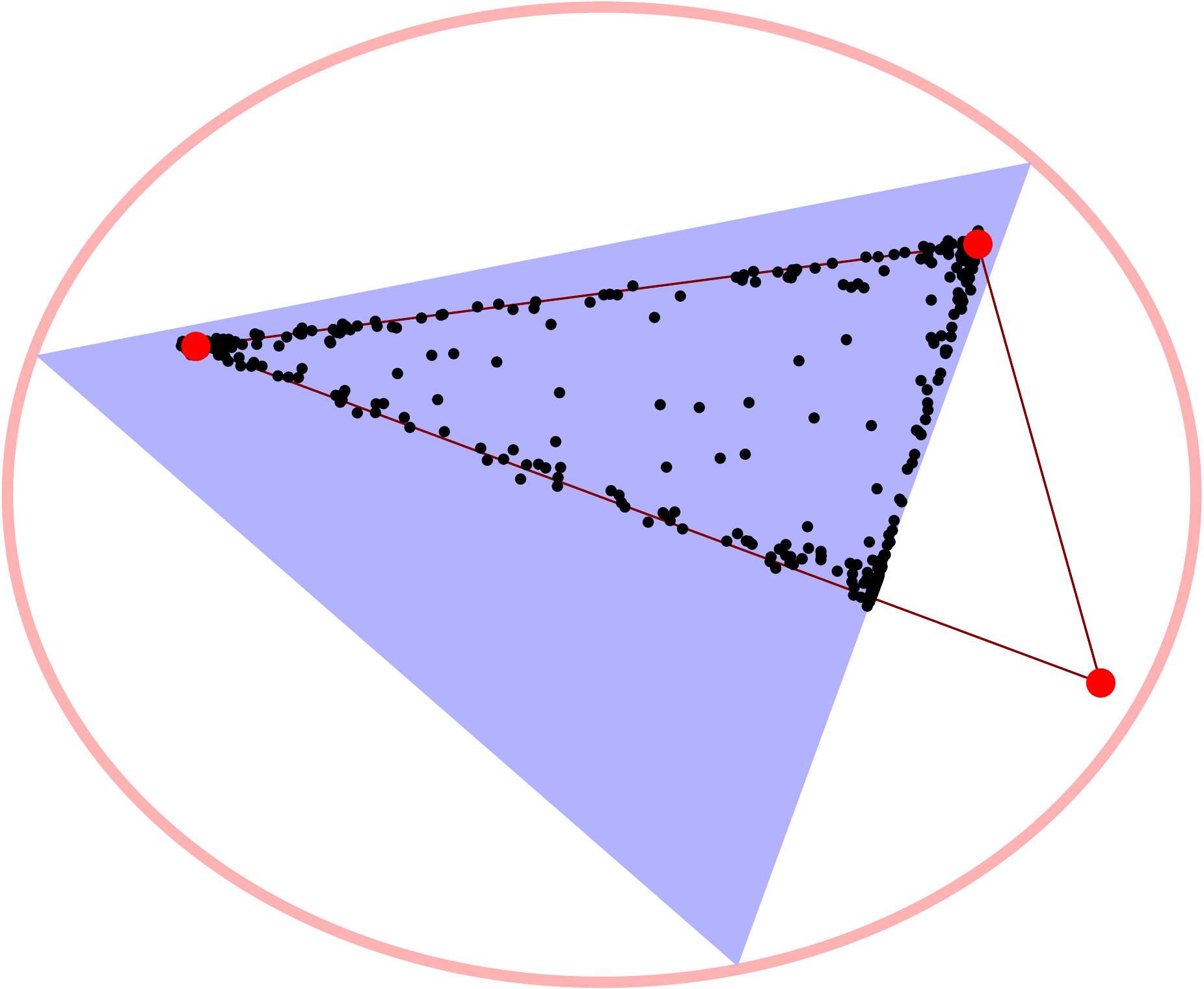}
        \caption{Simplex $\hat \beta$}
        \label{subfigure:exact_space}
    \end{subfigure}%
    \hspace{10pt}
    \begin{subfigure}[t]{0.45\linewidth}
        \centering
        \includegraphics[width=0.9\linewidth]{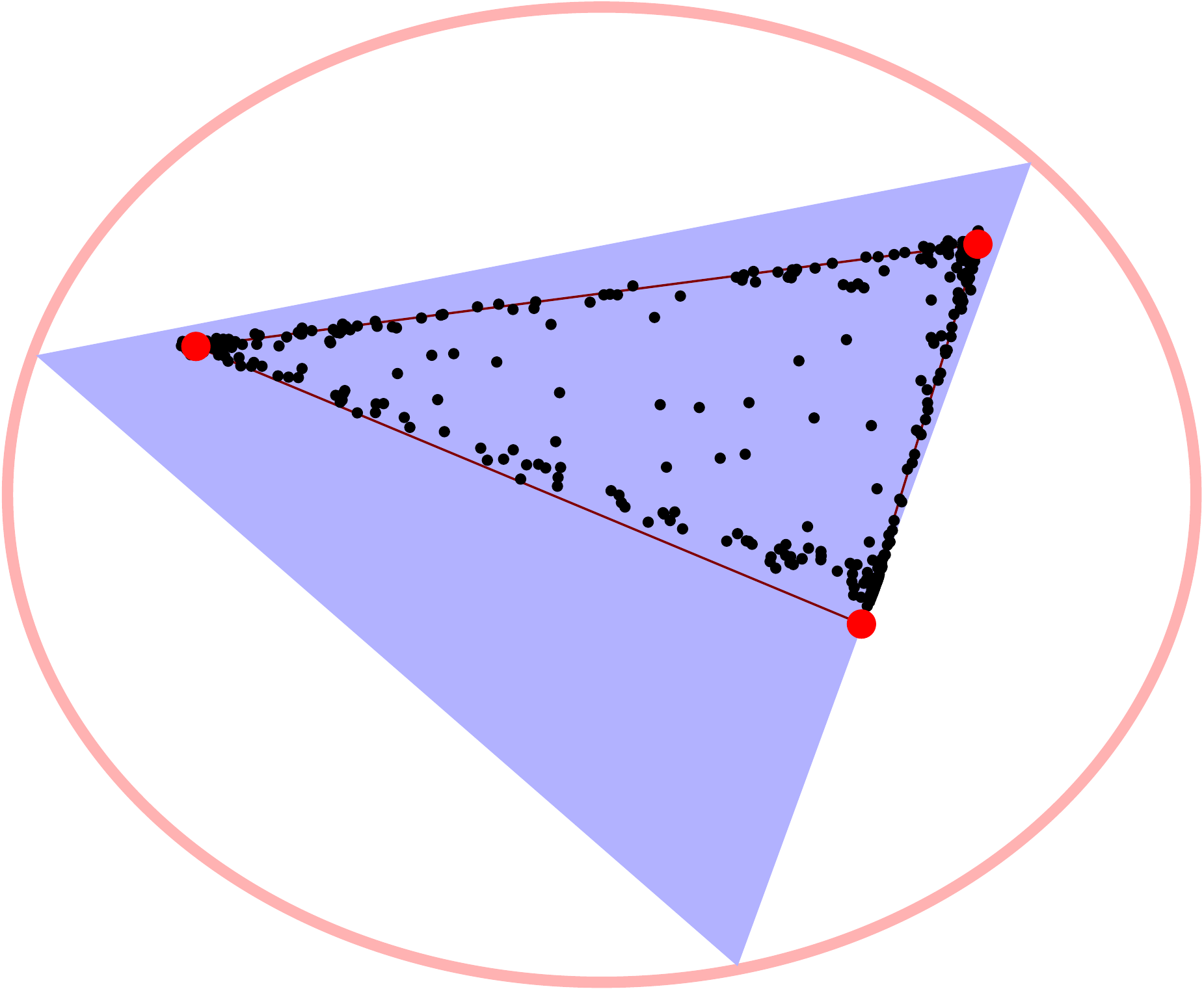}
        \caption{Simplex $\mt{Proj}_{\Delta^V} (\hat \beta)$}
        \label{subfigure:relaxed_space}
    \end{subfigure}%
    \centering
    \caption{Visualization of the proof for Lemma~\ref{lemma:sol_space}. The set $\Gamma$ corresponds to the blue triangle, which represents the feasible set of the problem \eqref{eq:geom_topic_obj}. The red circle represents the feasible set of the relaxed problem \eqref{eq:geom_topic_relaxed}. Given a potential solution $\widehat \beta$ for \eqref{eq:geom_topic_relaxed}, we can always argue that the projection of $\widehat \beta$, namely $\mt{Proj}_{\Delta^V}(\widehat \beta)$, is a better solution to \eqref{eq:geom_topic_relaxed} as illustrated in Panel~\ref{subfigure:relaxed_space}.  }
    \label{fig:lemma_pic}
\end{figure}

Intuitively, Lemma~\ref{lemma:sol_space} tells us that we cannot have the solution outside of the blue triangle in Figure~\ref{fig:lemma_pic}. If there was a solution outside of the triangle (Figure~\ref{subfigure:exact_space}), we could find the projection (Figure~\ref{subfigure:relaxed_space}) onto the word simplex (blue triangle) that still satisfies the constraint yet has a smaller volume, which is a contradiction.

\begin{proof}
    We prove this statement by contradiction. Suppose $\widehat \gamma \not \in \Gamma$. Then, as $\widehat \gamma$ is an optimal solution to the problem \eqref{eq:geom_topic_relaxed}, we have that $\theta = (WE_K)\widehat{\gamma} \in \mathbb{D}^{M\times K}$. Furthermore, since $W \in \Delta^V$ and $E_K$ is obtained from PCA, we have that $E_K^T \textbf{1}_V = 0$. Thus, it follows that
    \begin{align*}
        \widehat \beta \textbf{1}_V & = \overline{w}\textbf{1}_V  + (\widehat\gamma^{-1} - \overline{w}E_K)E_K^T \textbf{1}_V \\
        & = \textbf{1}_K - (\widehat\gamma^{-1} - \overline{w}E_K) \textbf{0}_V = \textbf{1}_K
    \end{align*}
    Therefore, the only constraint that $\widehat \gamma$ could possibly violate is non-negativity of $\widehat\beta$. Let $\mt{Proj}_{\Delta^V} (\widehat \beta)$ be the projection of $\widehat\beta$ onto the simplex $\Delta^V$ and let $\widehat{\gamma}_{\mt{proj}} = \mt{Proj}_{\Delta^V} (\widehat\beta) E_K$. Then, $\widehat{\gamma}_{\mt{proj}} \in \Gamma$ satisfies all the constraints in the optimization problem \eqref{eq:geom_topic_relaxed}, but we also have that
    \begin{equation*}
        -\log (\det \widehat{\gamma}_{\mt{proj}}\widehat{\gamma}_{\mt{proj}}^T) < -\log (\det \widehat \gamma \widehat\gamma^T)
    \end{equation*}
    since the volume of $ \mt{Proj}_{\Delta^V} (\widehat \beta)$ is smaller than that of $\widehat \beta$. This is a contradiction as $\widehat \gamma$ is the optimal solution to the problem \eqref{eq:geom_topic_relaxed}. Thus, it follows that $\widehat \gamma \in \Gamma$.
\end{proof}
We now state the sufficiently scattered assumption from \citet{huang2016anchor}.

\textbf{Assumption 1:} (\textit{sufficiently scattered condition} (\citet{huang2016anchor})) Let cone$(\beta)^*=\{x: \beta x\geq 0\}$ be the polyhedral cone of $\beta$ and $S=\{x: \norm{x}_2 \leq 1^T x\}$ be the second order cone. Matrix $\beta$ is called sufficiently scattered if it satisfies:

1) cone$(\beta)^* \subset S$

2) cone$(\beta)^* \cap bd(S) = \{a e_k: a \geq 0, k=1, \cdots, K\}$, where $bdS$ denotes the boundary of $S$.

The sufficiently scattered assumption can be interpreted as an assumption that we observe a sufficient number of documents on the faces of the topic simplex. In real-world topic model applications, such an assumption is not unreasonable since there are usually documents in the corpora having sparse representations.

\begin{proposition}
\label{thm:mvtm_suff_scatter}
Let $\gamma_*$ be the optimal solution to the problem \eqref{eq:geom_topic_relaxed} and $\beta_* =\overline{w} + (\gamma_*^{-1} - \overline{w}E_K)E_K^T$ be the corresponding topic matrix. If the true topic matrix $\beta$ is sufficiently scattered and rank$(\widetilde{W})=K$, then $\beta_* = \beta \Pi$, where $\Pi$ is a permutation matrix.
\end{proposition}
The proof structure is similar to the one in  \citet{huang2016anchor}, and we include here for completeness.
\begin{proof}
    Given a corpus $W \in \mathbb{D}^{M\times K}$, let $\beta\in \mathbb{D}^{K\times V}$ be the true topic-word matrix. Suppose rank$(W)=K$ and $\beta$ is sufficiently scattered. Let $\gamma_*$ be the solution to the problem \eqref{eq:geom_topic_relaxed}. Then, by Lemma~\ref{lemma:sol_space}, we have that $\gamma_* \in \Gamma$. Furthermore, since rank$(W) = K$, we have that rank$(\beta_*) = K$ as $W = \theta \beta_*$ where $\theta = (WE_K)\gamma_*$. It also follows that rank$(\beta)=K$ due to the constraint $W = \theta \beta$. Therefore, $|\det \beta|$ and $|\det \beta_*|$ are strictly positive. In other words, we cannot have a trivial solution to \eqref{eq:geom_topic_relaxed} as the objective is bounded. As $\beta$ and $\beta_*$ are full row rank, there exists an invertible matrix $Z\in \mathbb{R}^{K\times K}$ such that $\beta_* = Z \beta $. Also, as $\gamma_* \in \Gamma$, it follows that $\beta_* = Z \beta \geq 0$ and
    % \begin{align*}
    %     Z \beta \geq 0 \qquad
    %     & Z \beta \textbf{1}_V = Z \textbf{1}_K = \textbf{1}_K
    % \end{align*}
    \begin{align*}
        & \beta_* \textbf{1}_V = Z\beta \textbf{1}_V = \textbf{1}_K \\
        & \quad \Rightarrow Z \textbf{1}_K = \textbf{1}_K \\
    \end{align*}
    The inequality constraint $Z\beta \geq 0$ tells us that rows of $Z$ are contained in cone$(\beta)^*$. As $\beta$ is sufficiently scattered, it follows that
    \begin{equation}
    \label{eq:suff_cone}
        Z \big[k, :\big] \in \mt{cone}(\beta)^* \subset S
    \end{equation}
    by the first condition of (A1). Then, by the definition of the second order cone $S$, it follows that
    % , $\exists \Lambda$ such that $\beta(\Lambda, :)$ is a nonsingular diagonal matrix, and we have that
    % \begin{align*}
    %     Z \big[:, \Lambda\big] \beta\big[\Lambda, :\big]  = \beta_\omega \big[\Lambda, :\big] >0
    % \end{align*}
    % as $\beta\big[ \Lambda, : \big]$ is a nonnegative diagonal matrix, we have that $ Z \big[:, \Lambda\big] \beta\big[\Lambda, :\big] > 0 \Leftrightarrow Z > 0$. We also have that
    % \begin{align*}
    %     & \beta \textbf{1}_V = \beta_\omega \textbf{1}_V = \textbf{1}_K \\
    %     & \quad \Rightarrow Z \textbf{1}_K = \textbf{1}_K \\
    % \end{align*}
    % In other words, $Z \in \mathbb{D}^{K\times K}$. Thus, we have the following inequlaity
    \begin{equation}
    \label{eq:z_perm}
        \begin{aligned}
            & |\det Z|  = |\det Z^T| \leq \prod_{k=1}^K \norm{Z\big[k, :\big]}_2 \\
            % & \qquad \leq \prod_{k=1}^K\norm{Z\big[k, : \big]}_1 =
            & \qquad \leq \prod_{k=1}^K Z\big[ k, :\big] \textbf{1} = 1
        \end{aligned}
    \end{equation}
    The first inequality comes from the Hadamard inequality, which states that the equality holds if and only if the vectors $Z\big[ k,:\big]$'s are orthogonal to each other. The second inequality holds when $\norm{Z[k,:]}_2 = Z[k,:] \textbf{1}_K \mt{ } \forall k=1\cdots K$. In other words, when $Z[k,:] \in \mt{bd}(S) \mt{ } \forall k$. Then, together with \eqref{eq:suff_cone}, it follows that
    \begin{equation}
    \label{eq:suff_second_cond}
    \begin{aligned}
        Z[k,:] & \in  \mt{ cone}(\beta)^* \cap \mt{bd} S \\
        & = \{\lambda \textbf{e}_k| \lambda \geq 0, k=1, \cdots, K\}
    \end{aligned}
    \end{equation}
    Thus, it follows that the $|\det Z|$ achieves its maximum at 1, when $Z\in \mt{cone}(\beta)^* \cap \mt{bd} S$ sums to one and is an orthogonal matrix, i.e. when $Z$ is a permutation matrix.

    Furthermore, since $\gamma_* = (\beta_* E_K)^{-1} = Z^{-1}(\beta E_K)^{-1} = Z^{-1} \gamma$,
    we have that
    \begin{align*}
        - \det (\gamma_* \gamma_*^T) & = - \det (Z^{-1} \gamma \gamma^T (Z^{-1})^T)  \\
        & = - |\det Z^{-1}| \det (\gamma \gamma^T) |\det Z^{-1}| \\
        & = - |\det Z|^{-2} \det (\gamma \gamma^T) \\
        & \geq - |\det (\gamma \gamma^T)|
    \end{align*}
    where the equality holds when $|\det Z| = 1$. In other words, the minimum is achieved when $Z$ is a permutation matrix. Therefore, our solution $\gamma_*$ to the problem \eqref{eq:geom_topic_relaxed} and the corresponding $\beta_*$ are equal to the true topic-word matrix up to permutation.
\end{proof}

\textbf{Assumption 2} (\textit{Separability assumption} from \citet{arora2013practical}) There exists a set of indices $\Lambda = \{i_1, \cdots, i_K\}$ such that $\beta(\Lambda, :) =$ Diag(c), where $c\in \mathbb{R}_+^K$.

The separability assumption, also known as the anchor-word assumption, states that every topic $k$ has a unique word $w_k$ that only shows up in topic $k$. These words are also referred to as the anchor words as introduced in \citet{arora2013practical}.

\textbf{Remark:} The identifiability statement in Proposition~\ref{thm:mvtm_suff_scatter} holds true under the separability assumption as well, as the sufficiently scattered assumption is a weaker version of the separability assumption.
%%%%%%%%%%%%%%%%%%%%%%%%%%%%%%%%%%%%%%%%%%%%%%%

\subsection{Augmented Lagrangian Formulation}

With  $\mu \geq 0$, we work with the following augmented Lagrangian version of the constrained optimization problem \eqref{eq:geom_topic_relaxed}
\begin{equation}
\label{eq:geom_topic_relaxed_lag}
    \begin{aligned}
        \widehat{\gamma} &= \argmin_\gamma -\log|\det(\gamma \gamma^T)| + \mu \norm{\widetilde{W} \gamma}_h\\
        & \mt{s.t.}  \quad  \gamma \textbf{1}_K = (\widetilde{W}^T \widetilde{W})^{-1} \widetilde{W}^T \textbf{1}_V \quad \sigma_{min}(\gamma ) \geq \zeta \\
    \end{aligned}
\end{equation}
where $\widetilde{W} = WE_K$, $\zeta = R^{-1}$, and $\norm{X}_h = \sum_{i,j} \max(-X_{i,j}, 0)$ is a hinge loss that captures the non-negativity constraint on $\theta$. Furthermore, the linear constraint is converted to $\gamma \textbf{1}_K = (\widetilde{W}^T \widetilde{W})^{-1} \widetilde{W}^T \textbf{1}_V $, which is the same constraint as $\widetilde{W}\gamma \textbf{1}_K = \textbf{1}_V$. For simplicity, we define $\textbf{a} = (\widetilde{W}^T \widetilde{W})^{-1} \widetilde{W}^T \textbf{1}_V $.

The Lagrangian objective function in \eqref{eq:geom_topic_relaxed_lag} can be written as
\begin{equation}
\label{eq:topic_vol_obj}
    \begin{aligned}
        f(\gamma)  = &  -\log |\det(\gamma \gamma^T)| + \mu \norm{\widetilde{W}\gamma}_h \\
        & \quad + \mathds{1}(\sigma_{min}(\gamma) > \zeta)  \mt{ s.t. }\gamma \textbf{1}_K = \textbf{a}
    \end{aligned}
\end{equation}
Introducing the auxiliary optimization variables $V_1 \in \mathbb{R}^{n\times k}$ and $V_2 \in \mathbb{R}^{k\times k}$, we reformulate \eqref{eq:geom_topic_relaxed}
\begin{equation}
\label{eq:geom_topic_aux}
    \begin{aligned}
        \hat \gamma & =  \argmin_{\gamma, V_1, V_2} \Big \{
         -\log |\det \gamma\gamma^T| + \mu \norm{V_1}_h + \\
        & \qquad \qquad \qquad  + \mathds{1}(\sigma_{min}(V_2) > \zeta) \Big \} \\
        &  \qquad \mt{s.t.} \quad V_1 = \widetilde{W} \gamma \quad \gamma = V_2 \quad \gamma \textbf{1}_K = \textbf{a}
    \end{aligned}
\end{equation}
For a penalty parameter $\rho > 0$ and Lagrange multiplier matrix $\Lambda \in \mathbb{R}^{n \times k}$, we consider the augmented Lagrangian of this problem
\begin{equation}
\label{eq:topic_logdet_lag}
    \begin{aligned}
        & \mathcal{L}(\gamma, V_1, V_2, \Lambda_1, \Lambda_2) \\
        & = -\log |\det \gamma \gamma^T | + \mu \norm{V_1}_h +  \mathds{1}(\sigma_{min}(V_2) > \zeta) \\
        & \quad + \frac{\rho}{2} \norm{\widetilde{W} \gamma - V_1}_F^2 + \langle \Lambda_1, \widetilde{W} \gamma - V_1 \rangle \\
        & \quad + \frac{\rho}{2} \norm{\gamma - V_2}_F^2 + \langle \Lambda_2,  \gamma - V_2 \rangle \quad \mt{ s.t. } \gamma \textbf{1}_K =\textbf{a}
    \end{aligned}
\end{equation}
This function can be minimized using an iterative ADMM update scheme on the arguments $\gamma$, $V_1$, $V_2$, $\Lambda_1$, and $\Lambda_2$. The update for $V_1$ and $V_2$ can be accomplished by standard proximal operators that implement soft-thresholding and a projection. Furthermore, the $\gamma$-update can be derived in a closed form by solving a quadratic equation in its singular values. The details of the ADMM updates are included in the supplement.
% We can derive the closed-for update for $\gamma^{t+1}$ as well, since it is a convex problem with a linear constraint.
First, consider the $\gamma$-subproblem without the linear constraint $\gamma \textbf{1} = \textbf{a}$. Then, as derived in the supplement, the resulting update equation for $\gamma$ is
\begin{equation}
\label{eq:def_gamma_plus}
\begin{aligned}
    \gamma^+ & = \argmin_{\gamma \in \mathbb{R}^{k\times k}} \Big \{ -\log |\det \gamma^T \gamma| + \frac{\rho}{2}\norm{C^{1/2} (\gamma - A)}_F^2 \Big \} \\
    & = U \widehat{D} W^T
\end{aligned}
\end{equation}
where $\widehat{D}$ is defined in the supplement. Using $\gamma_+$, we obtain a closed-form solution to the $\gamma$ sub-problem in \eqref{eq:topic_logdet_lag} as follows
\begin{align*}
    \gamma^{t+1} = \gamma_+ - (\gamma_+ \textbf{1} - \textbf{a})(\textbf{1}^T C^{-1} \textbf{1})^{-1} \textbf{1}^T C^{-1}
\end{align*}
This solution to the linear constrained problem can be easily derived as a stationary point of the convex function that is minimized in \eqref{eq:def_gamma_plus}. Note that, by construction, $\gamma^{t+1} \textbf{1} = \textbf{a}$.
\begin{algorithm}
\caption{Minimum volume topic modeling}
  \SetAlgoLined
  \KwIn{W, $E_K$, $\gamma^0$, $\rho > 0$, $\mu>0$}
  \KwOut{$\hat{\beta}$}
  Initialize $V_2^0 = \gamma^0$, $V_1 = \widetilde{W} \gamma^0$, $\Lambda_1^0 = \m{0}$, $\Lambda_2^0 = \m{0}$ \;
  Calculate $C = I + \widetilde{W}^T \widetilde{W}$ \;
  Calculate the projected documents $\widetilde{W}$ \;
  \While{not converged}{
    $V_1^{t+1} = \mt{Prox}_{\norm{\cdot}_{h, \mu} / \rho} \left( \frac{\rho \widetilde{W} \gamma^t + \Lambda_1^t}{\rho} \right)$\\
    $V_2^{t+1} = \mt{Proj}_{G_R}\left( \frac{\rho \gamma^t + \Lambda_2^t}{\rho}\right)$\\
    $\gamma^{t+1} = \gamma_+ - (\gamma_+ \textbf{1} - \textbf{a})(\textbf{1}^T C^{-1} \textbf{1})^{-1} \textbf{1}^T C^{-1}$ \\
    $\quad$ where $\gamma_+$ is defined in \eqref{eq:def_gamma_plus} \\
    $\Lambda_1^{t+1} = \Lambda_1^k + \rho(\widetilde{W} \gamma^{t+1} - V_1^{t+1})$ \\
    $\Lambda_2^{t+1} = \Lambda_2^k + \rho(\gamma^{t+1} - V_2^{t+1})$ \\
  }
\label{alg:topic_vol_admm}
\end{algorithm}

In the nonnegative matrix factorization literature, \citet{liu2017large} used a large-cone penalty that constrains either the volume or the pairwise angles of the simplex vertices. However, this does not impose a sum-to-one constraint on the topics, and the optimization is performed over $\beta$. Furthermore, our formulation has an advantage over the problem in \citet{liu2017large} as we directly work with the latent topic proportions $\theta$. This is possible in our formulation as we decoupled $\beta$ from $\theta$ using the ADMM mechanism.

\subsection{Convergence}

The following proposition shows that Algorithm~\ref{alg:topic_vol_admm} converges to a stationary point of \eqref{eq:topic_vol_obj}.

\begin{proposition}
\label{thm:admm_convergence}
    For any limit point $(\gamma^*, V_1^*, V_2^*, \Lambda_1^*,\Lambda_2^*)$ of Algorithm~\ref{alg:topic_vol_admm}, $\gamma^*$ is also a stationary point of \eqref{eq:topic_vol_obj}.
\end{proposition}
This follows by applying a standard convergence proof of the ADMM algorithm (Algorithm~\ref{alg:topic_vol_admm}) based on the KKT condition. The proposition states that our ADMM formulation converges to a stationary point. However, while the unconstrained objective function in \eqref{eq:topic_vol_obj} is convex, the constraint on the minimum singular value makes the constrained optimization function non-convex. Thus, our algorithm is only guaranteed to converge to a stationary point of \eqref{eq:topic_vol_obj}.

\begin{figure}[thb!]
    \centering
    \begin{subfigure}[t]{0.48\linewidth}
        \centering
        \includegraphics[width=1.0\linewidth]{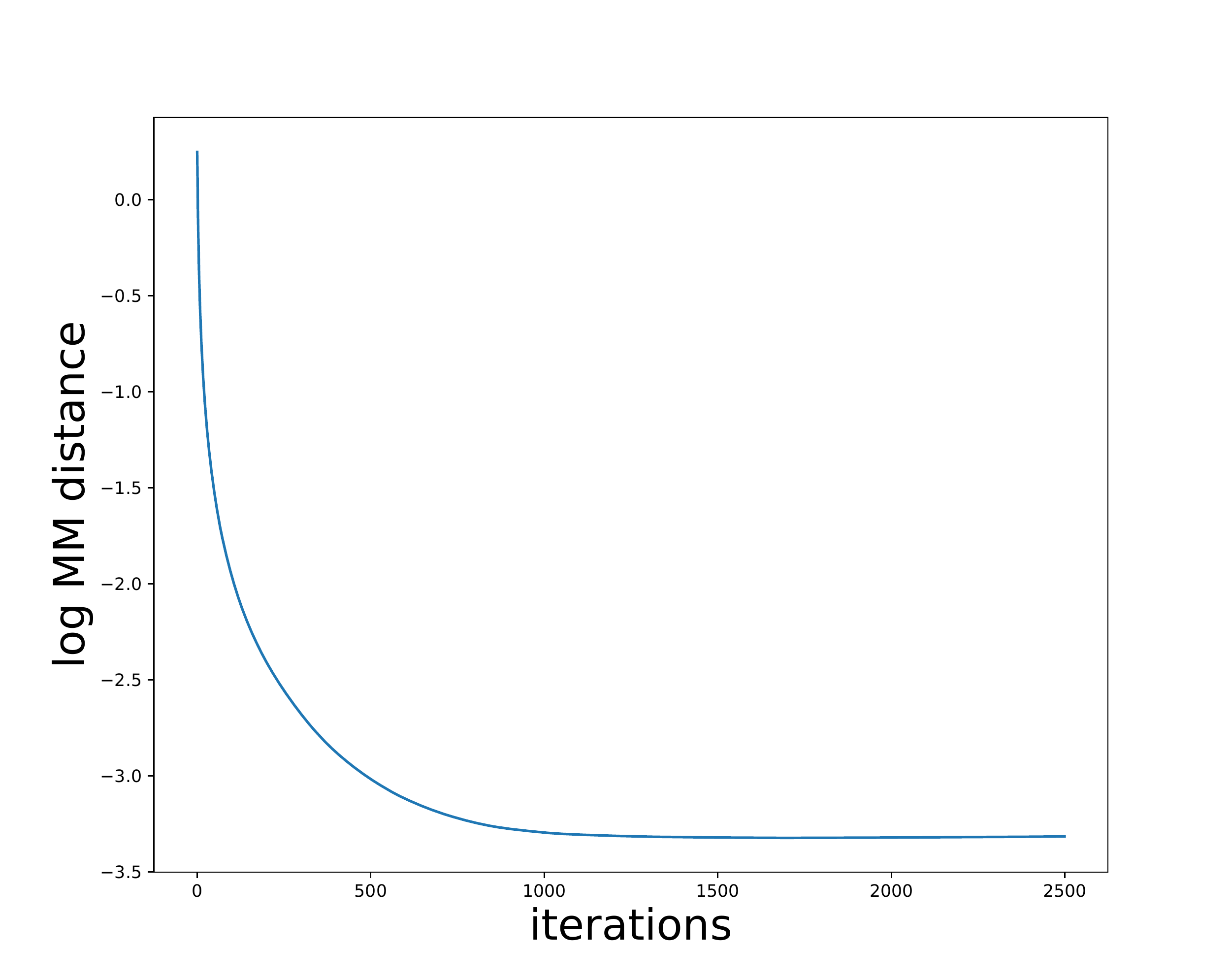}
        \caption{Frobenius Norm}
    \end{subfigure}%
    \begin{subfigure}[t]{0.48\linewidth}
        \centering
        \includegraphics[width=1.0\linewidth]{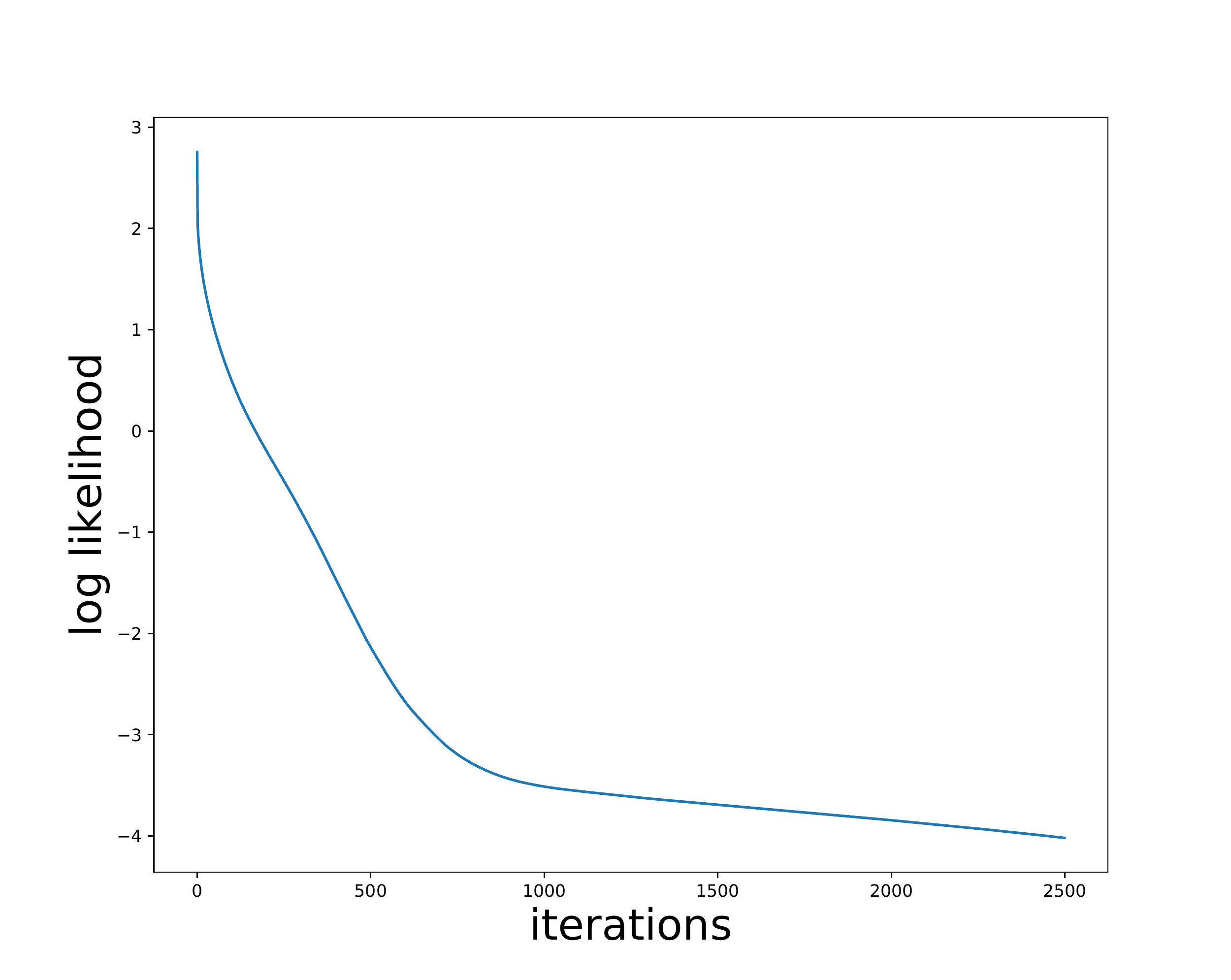}
        \caption{Objective}
    \end{subfigure}%
    \caption{Experimental runs using Algorithm~\ref{alg:topic_vol_admm}. The data was simulated from an LDA model with $\alpha = 0.1, \eta = 0.1, V=1200, K=3, M=1000, N_m=1000$. The algorithm was initialized with $\gamma$ equal to the identity matrix. The left panel shows the relative Frobenius error between the iterates $\gamma^t$ and the true $\gamma$. The right panel shows the convergence in terms of the objective values.}
    \label{fig:topic_vol_convergence}
\end{figure}

Figure~\ref{fig:topic_vol_convergence} demonstrates the convergence of our algorithm with synthetic data generated from an LDA model with parameters $\alpha = 0.1, \eta = 0.1, V=1200, K=3, M=1000$, and $N_m=1000$.

\section{Performance Comparison}
\begin{figure}
    \centering
    \includegraphics[width=0.65\linewidth]{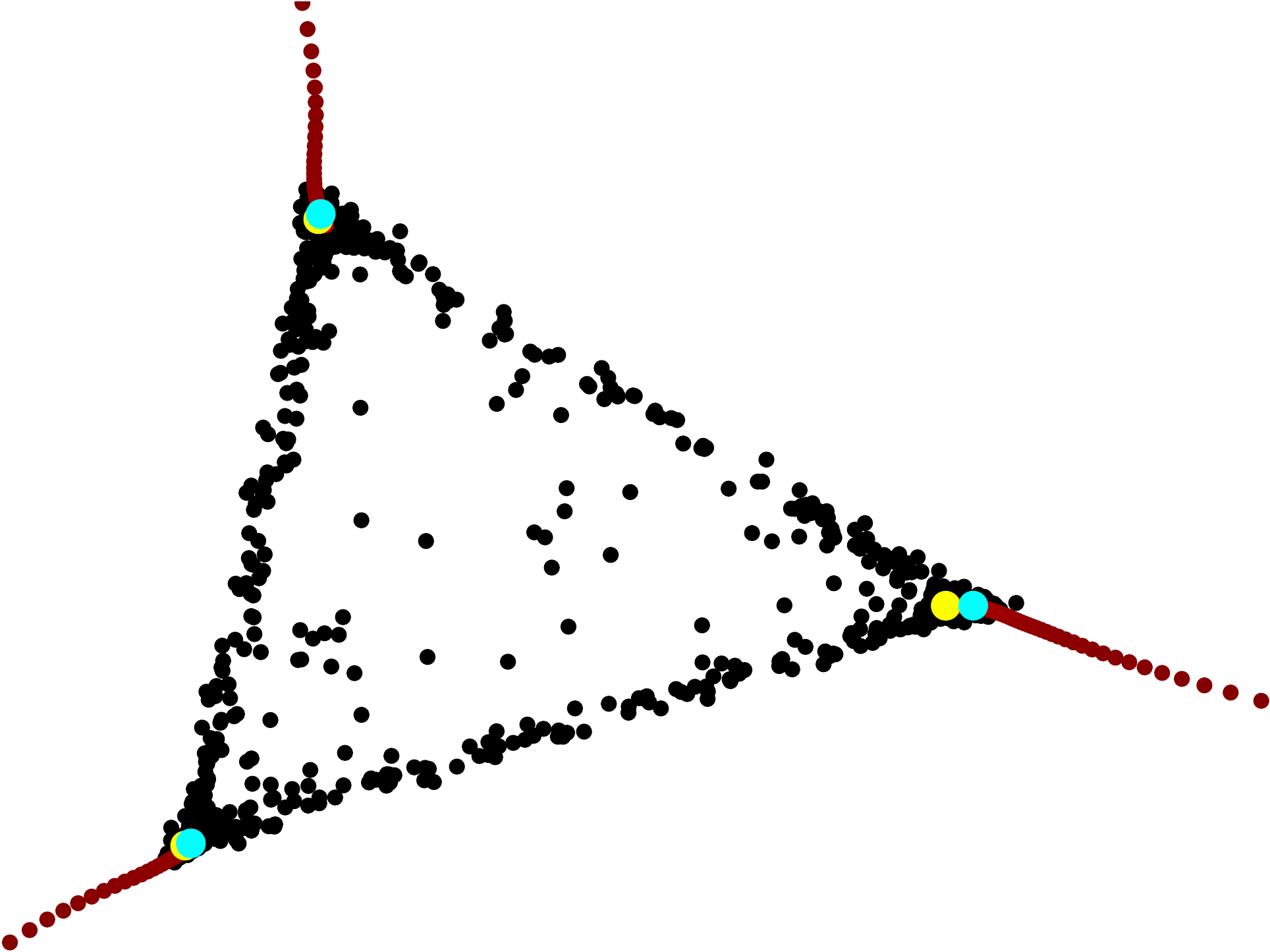}
    \caption{Visualization of Minimum Volume Topic Modeling (MVTM) with the observed documents in black, optimization path of MVTM in the gradient of red (dark red = beginning, light red = end), and the final estimate in yellow. The ground-truth topic vertices are plotted in cyan. The Dirichlet parameter for the topic proportion was set at $\alpha=0.1$, and MVTM was initialized at the identity matrix.}
    \label{fig:mvtm_toy_example_sep}
\end{figure}

\begin{figure}[h!]
  \centering
  \begin{subfigure}[t]{0.48\linewidth}
      \centering
      \includegraphics[width=\linewidth]{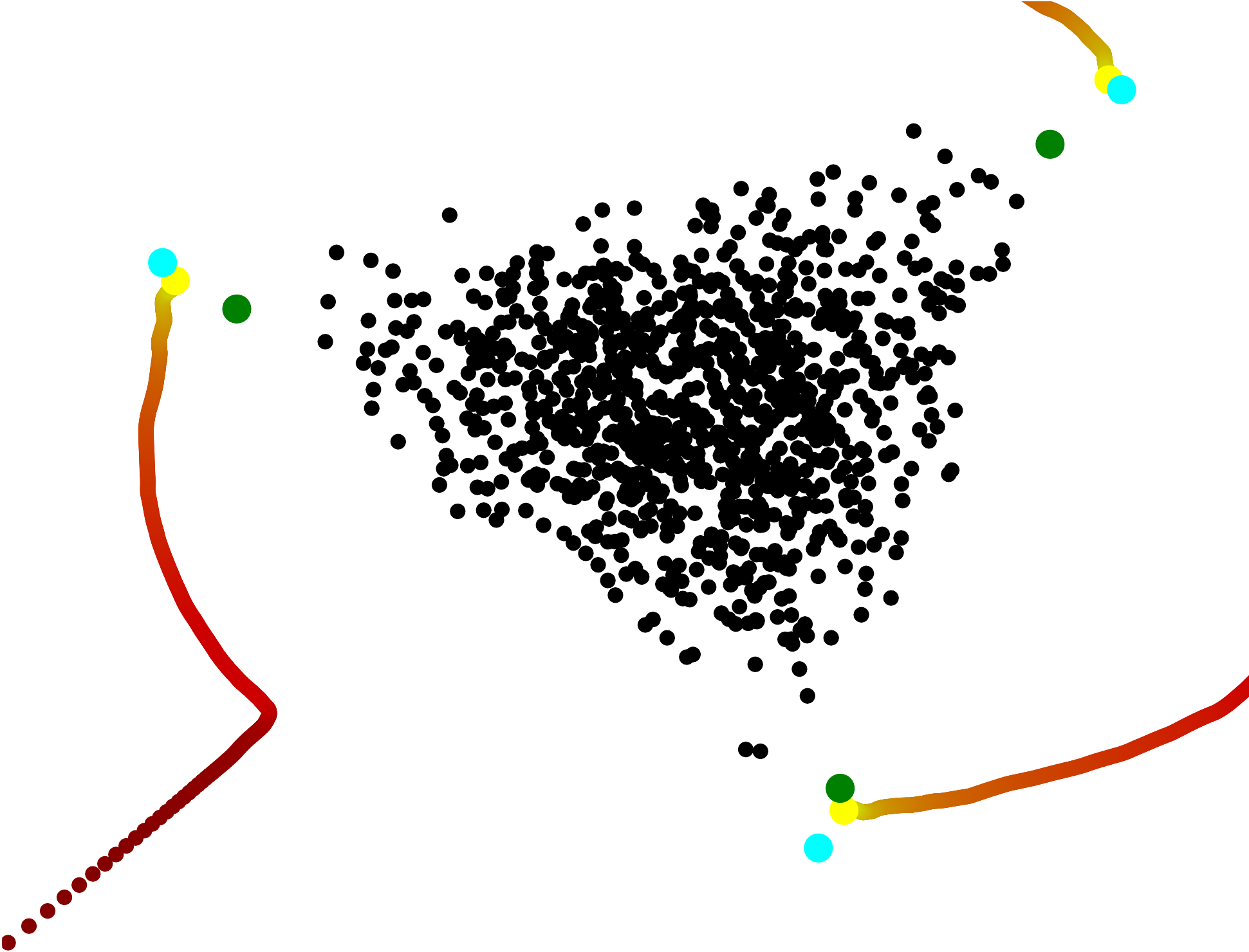}
      \caption{$\alpha =3$}
  \end{subfigure}%
  ~
  \begin{subfigure}[t]{0.48\linewidth}
      \centering
      \includegraphics[width=\linewidth]{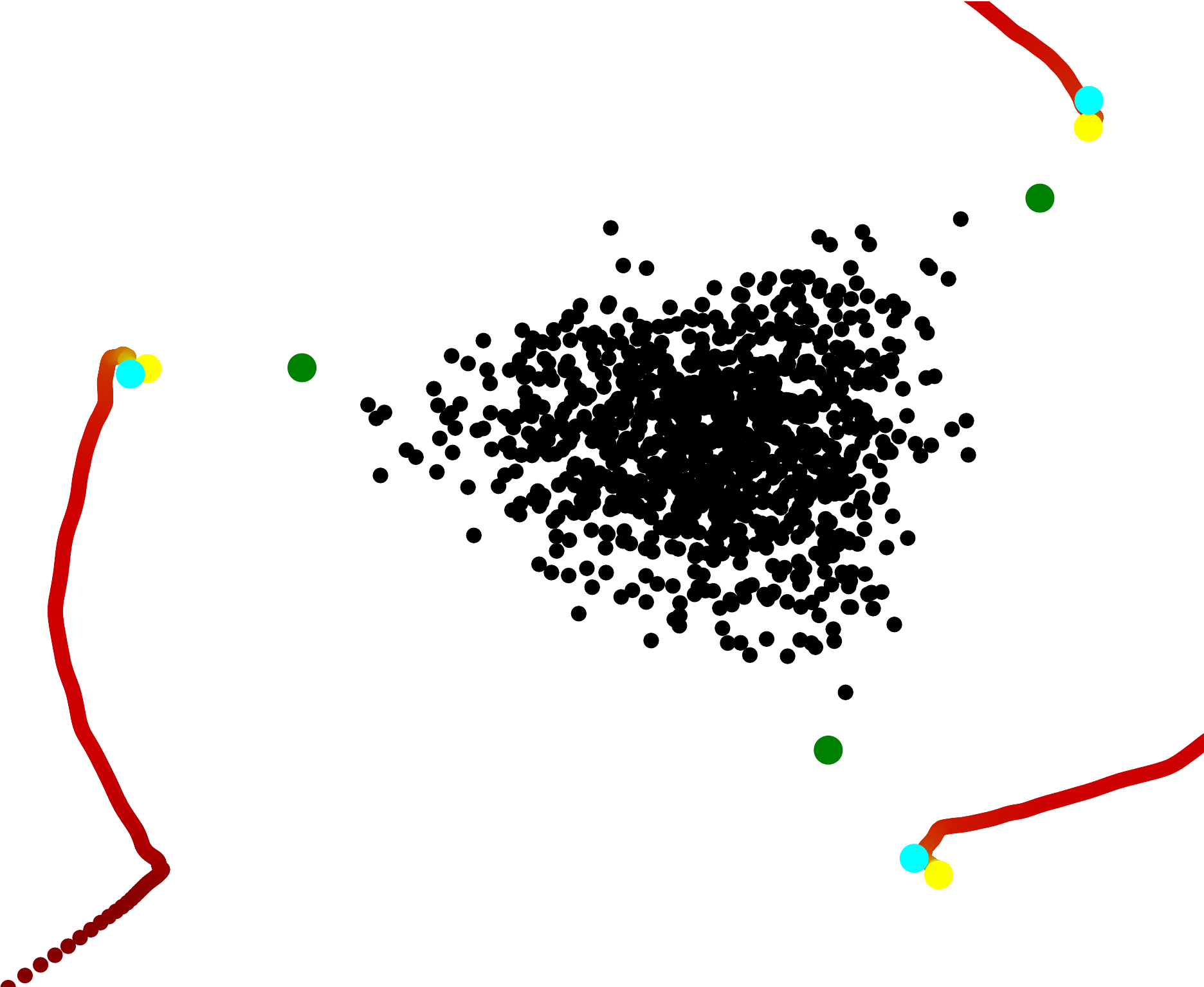}
      \caption{$\alpha = 5$}
      \label{subfig:sim_alpha5}
  \end{subfigure}%
    \caption{Visualization of the proposed MVTM algorithm with the observed documents in black, optimization path of MVTM in the gradient of red (dark red = beginning, light red = end), and the final estimate in yellow under different values of $\alpha$. The ground-truth topic vertices are plotted in cyan, and the final estimate of GDM is plotted in green for comparison. MVTM was initialized at the identity matrix.}
    \label{fig:mvtm_toy_example}
\end{figure}

To demonstrate the performance of the proposed minimum volume topic model (MVTM) estimation algorithm (Algorithm~\ref{alg:topic_vol_admm}), we generate the LDA data with the parameters $\eta = 0.1, V=1200, K=3, M=1000, N_m=1000$ with varying $\alpha$, which is Dirichlet hyperparameter for the topic proportion $\theta$. For ease of visualization, the first two dimensions of the projected documents and the estimated topics are used. The first scenario ($\alpha = 0.1$) in Figure~\ref{fig:mvtm_toy_example_sep} shows the performance of our algorithm is comparable to the vertex based method GDM (\citet{yurochkin_long_16}), when there are plenty of observed documents around the vertices. While there is no anchor word in the generated dataset, we observe enough documents around the vertices. In other words, the separability assumption is slightly violated.
% However, as we increase $\alpha$, i.e. when the documents are well mixed, MVTM recovers the correct topics with appropriate parameters for the hinge loss.
With higher values of $\alpha$, however, Figure~\ref{fig:mvtm_toy_example} shows the advantages of our method, denoted as MVTM.
Note that the higher values of $\alpha$ correspond to the situation where the sufficiently scattered condition is satisfied, but the separability condition is violated. Thus, we can see the vertex based method (GDM) starts to suffer in the oracle performance. In contrast, with an appropriate choice of $\mu$ for the hinge loss, our method recovers the correct topics even for the well mixed scenario where $\alpha = 5$. Figure~\ref{subfig:sim_alpha5} shows that there is a kink in the optimization path, where MVTM is finding the right orientation of the true simplex. Furthermore, there is a lack of loops in the optimization path, illustrating the identifiability of MVTM.

Lastly, we explore the asymptotic behavior by varying document lengths $N_m$ with $M=1000$, $K=5$, $V=1200$, $\eta=0.1$, $\alpha = 0.1$ and 100 held-out documents. MVTM is all initialized at the identity matrix, and VEM had 10 restarts as the objective for the variational method is nonconvex.

\begin{figure}[h] \centering

\begin{subfigure}[t]{0.7\linewidth} \centering
\includegraphics[width=\linewidth]{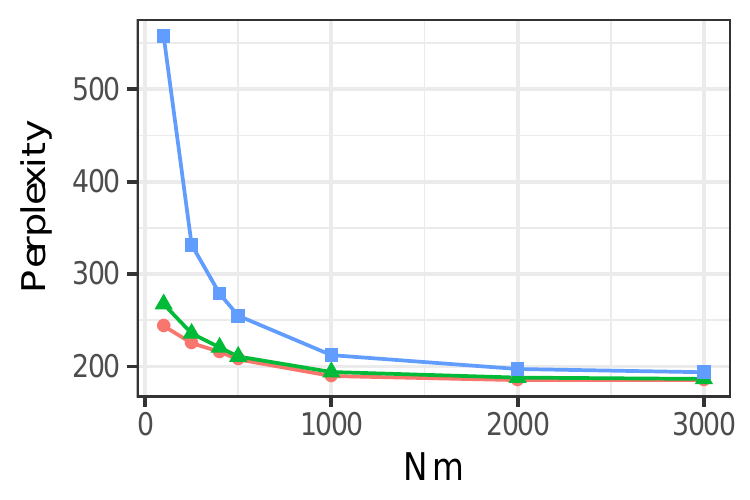}
\end{subfigure}

\begin{subfigure}[t]{0.7\linewidth} \centering
\includegraphics[width=\linewidth]{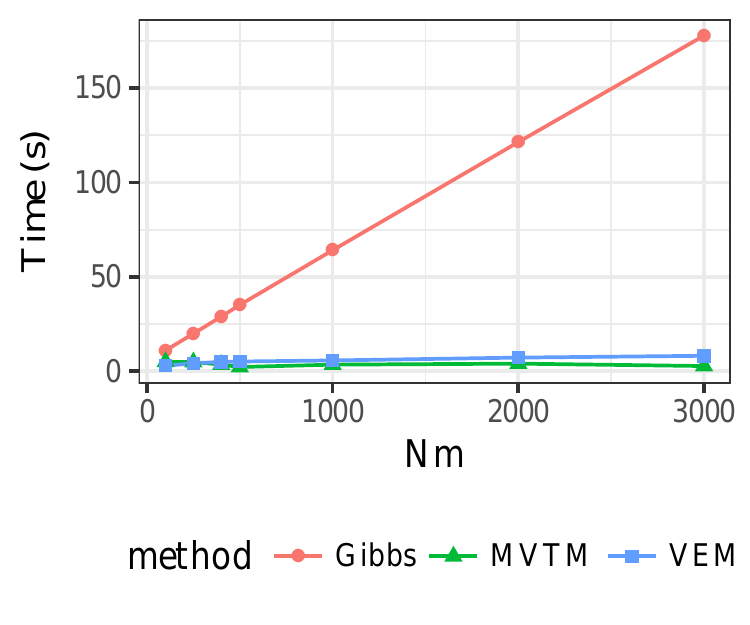}
\end{subfigure}
\vspace{-10pt}
\caption{Perplexity of the held-out data and the corresponding time complexity of each method at varying values of the number words per document $N_m$ with $M = 1000$, $K=5$, $V=1200$, $\eta=0.1$ and $\alpha=0.1$}
\label{fig:sim_Nm}
\end{figure}

Figure~\ref{fig:sim_Nm} tells us that 1) Gibbs sampling and MVTM have comparable performance in terms of perplexity, 2) MVTM and VEM both show the computational advantages over the Gibbs sampling method, and 3) VEM suffers from the statistical performance due to the nature of the non-convex objective function of VEM. Additional simulation results can be found in the supplement.

\subsection{NIPS dataset}
To illustrate the performance of MVTM on a real-world data, we apply our algorithm to NIPS dataset. We preprocess the raw data using a standard stop word list and filter the resulting data through a stemmer. After preprocessing, words that appeared more than 25 times across the whole corpus are retained. Then, we further remove the documents that have less than 10 words. The final dataset contained 4492 unique words and 1491 documents with mean document length of 1187. We compare our algorithm's performance to GDM and Gibbs sampling at K=5, 10, 15, and 20. The perplexity score is used to perform the comparison in Table~\ref{table:nips_perp}.
\begin{table}[h!]
\centering
\begin{tabular}{lclclclcl}
\hline
 & MVTM & GDM & RecoverKL & Gibbs  \\
\hline
K=5 & 1483 & 1602 & 1569 & 1336  \\
K=10 & 1387 & 1441 & 1507 & 1192   \\
K=15 & 1293 & 1344 & 1438 & 1109   \\
K=20 & 1273 & 1294 & 1574 & 1068   \\
% K=25 & 1230 & 1242 & *& 1023  \\
\hline
\end{tabular}
\caption{Perplexity score of the geometric algorithms and the Gibbs sampling for analyzing the NIPS dataset. The proposed algorithm MVTM is performing better than the vertex methods (GDM and RecoverKL) in terms of perplexity as it only requires the documents lie on the face of the topic simplex. GDM provides a similar performance to MVTM.}
\label{table:nips_perp}
\end{table}
The additional time comparison and top 10 words of top 10 learned topics for MVTM, GDM, and Gibbs sampling are provided in the supplement.
\section{Discussion}
This paper presents a new estimation procedure for LDA topic modeling based on the minimization of the volume of the topic simplex $\beta$. Such formulation can be thought of as an asymptotic estimation to the LDA model. The proposed minimum volume topic model (MVTM) algorithm differs from moment-based methods including RecoverKL and the vertex based method such as the GDM. We proved the identifiability of MVTM under the sufficiently scattered assumption introduced in \citet{huang2016anchor}. When the sufficiently scattered assumption is satisfied and the separability assumption is violated, MVTM continues to perform well with an appropriate choice of the hinge loss parameter.

There are open questions on the statistical convergence of our estimator in terms of the document length and the number of documents. Such relationships have been explored in the work of \citet{tang2014understanding}, and it would be interesting to see if these could be applied to the proposed MVTM. The understanding of the statistical behavior of MVTM will provide us with the theoretical guidance on the choice the hinge loss parameter. Besides the theoretical questions, MVTM also has some potential modeling extensions. The immediate extension includes the nonparametric setting, where one would also estimate the number of topics $K$.

\subsubsection*{Acknowledgments}
This research was partially supported by grant ARO W911NF-15-1-0479.

\vfill
\pagebreak
\clearpage
% \nocite{*}
\bibliographystyle{apa}
\bibliography{biblio_combined}

\clearpage
\vfill
\pagebreak
\appendix

\section{Algorithm Analysis}
\label{appendix:admm}
\subsection{ADMM update derivation}
For completeness, we derive the ADMM steps of the problem in \eqref{eq:topic_logdet_lag}. Given current iterates $V_1^t, \gamma^t,$ and $\Lambda^t$,
\begin{equation}
\label{eq:v1_update}
\begin{aligned}
    V_1 ^{t+1} & = \argmin_{V_1 \in \mathbb{R}^{n\times k}} \Big \{ \mu \norm{V_1 }_h  + \frac{\rho}{2} \norm{\widetilde{W} \gamma^t - V_1 }_F^2 \\
    & \qquad \qquad \qquad \qquad \qquad + \langle \Lambda_1, \widetilde{W} \gamma^t - V_1  \rangle  \Big \} \\
    & = \argmin_{V\in \mathbb{R}^{n\times K}} \left \{ \frac{\mu}{\rho} \norm{V_1}_h + \frac{1}{2} \norm{V_1 - \frac{\rho \widetilde{W} \gamma^{t} + \Lambda_1^t}{\rho}}_F^2\right\} \\
    & = \mt{Soft-Threshold}_{\mu/\rho}\left( \frac{\rho \widetilde{W} \gamma^{t} + \Lambda_1^t}{\rho} \right)
\end{aligned}
\end{equation}
where we soft-threshold the matrix with the regularization parameter $\frac{\lambda}{\rho}$.
\begin{equation}
\label{eq:v2_update}
\begin{aligned}
    V_2^{t+1} & = \argmin_{V_2 \in \mathbb{R}^{k\times k}} \Big \{  \frac{\rho}{2} \norm{\gamma^t - V_2}_F^2 + \langle \Lambda_2,  \gamma^t - V_2 \rangle \\
    & \qquad \qquad \qquad \qquad + \mathds{1}(\lambda_{min}(V_2 V_2^T) \geq \frac{1}{R^2}) \Big\}\\
    & = \argmin_{V_2 \in \mathbb{R}^{k\times k}} \Big \{ \frac{1}{2} \norm{V_2 - \frac{\rho \gamma^t + \Lambda_2^t}{\rho}}_F^2  \\
    & \qquad \qquad \qquad \qquad \qquad + \mathds{1}(\sigma_{min}(V_2) \geq \frac{1}{R}) \Big\} \\
    & = \mt{Proj}_{G_R}\left(  \frac{\rho \gamma^t + \Lambda_2^t}{\rho}\right)
\end{aligned}
\end{equation}
where $G_R = \{ X\in \mathbb{R}^{n\times K} | \sigma_{min}(X) \geq \frac{1}{R}\}$ and $\mt{Proj}_{G_R}$ is the projection onto the set $G_R$.
\begin{equation}
\label{eq:gamma_problem}
\begin{aligned}
    \gamma^{t+1} & = \argmin_{\gamma \in \mathbb{R}^{k\times k}} \Big \{  -\log |\det \gamma \gamma^T |   +  \frac{\rho}{2} \norm{\widetilde{W} \gamma - V_1}_F^2 \\
    & \qquad \qquad \qquad + \langle \Lambda, \widetilde{W} \gamma - V_1 \rangle +  \frac{\rho}{2} \norm{\gamma - V_2}_F^2 \\
    & \qquad \qquad \qquad  + \langle \Lambda_2,  \gamma - V_2 \rangle \Big \} \quad \mt{s.t.} \quad \gamma \textbf{1}_K = \textbf{a} \\
    & = \argmin_{\gamma \in \mathbb{R}^{k\times k}} \Big \{ -\log |\det \gamma \gamma^T| + \frac{\rho}{2}\norm{C^{1/2} (\gamma - A)}_F^2 \Big \} \\
    & \qquad \qquad \mt{s.t.} \quad \gamma \textbf{1}_K
     = \textbf{a}
\end{aligned}
\end{equation}
where we have that
\begin{align*}
    A & = C^{-1}B^T = UD_AV^T\\
    B & = (V_1^{t+1})^T \widetilde{W} + (V_2^{t+1})^T - \frac{(\Lambda_2)^t)^T}{\rho} - \frac{(\Lambda_1)^t)^T \widetilde{W}}{\rho} \\
    C & = I + \widetilde{W}^T \widetilde{W}
\end{align*}
We can derive the update for $\gamma^{t+1}$, as it is a convex problem with a linear constraint. First, consider the \eqref{eq:gamma_problem} without the linear constraint $\gamma \textbf{1} = \textbf{a}$. Then, we can rewrite the unconstrained $\gamma$-subproblem as

\begin{align*}
    & \gamma_+  = \argmin_{\gamma \in \mathbb{R}^{k\times k}} \Big \{ -\log (\det \gamma \gamma^T) + \frac{\rho}{2}\norm{C^{1/2} (\gamma - A)}_F^2 \Big \} \\
    &  = \argmin_{\gamma \in \mathbb{R}^{k\times k}} \Big \{ -\log (\det \gamma \gamma^T) + \frac{\rho}{2} \tr(\gamma^T C\gamma) \\
    & \qquad \qquad \qquad \qquad \qquad \qquad \qquad - \rho \tr(\gamma^T C A) \Big \} \\
    & = \argmin_{\gamma =UDV^T} \Big \{ -\log (\det \gamma \gamma^T) + \frac{\rho}{2} \tr(\gamma^T C\gamma) \\
    & \qquad \qquad \qquad \qquad \qquad \qquad \qquad - \rho \tr(\gamma^T C A)\Big \} \\
    & = \argmin_{\gamma =UDV^T} \Big \{ -\log (\det D^2) + \frac{\rho}{2} \tr(UD^2U^T C) \\
    & \qquad  \qquad \qquad \qquad \qquad \qquad - \rho \tr(UD_A DU^T C)\Big \} \\
    & = \argmin_{\gamma =UDV^T} \Big \{ -\sum_{i=1}^K 2\log |D_{ii}| + \frac{\rho}{2} \tr(ED^2) - \rho \tr(FD)\Big \} \\
    & = \argmin_{\gamma =UDV^T} \Big \{ -\sum_{i=1}^K 2\log |D_{ii}| + \frac{\rho}{2} E_{ii} D_{ii}^2 - \rho F_{ii} D_{ii}\Big \} \\
\end{align*}
where $E = U^T C U$ and $F = U^T C U D_A$. Then we can solve the above problem element by element. Looking at the $i$-th entry, we can take the derivative and set it to zero. That is
\begin{align*}
    & \frac{\partial}{\partial D_{ii}} \left(\log |D_{ii}| + \frac{\rho}{2} E_{ii} D_{ii}^2 - \rho F_{ii} D_{ii}\right ) = 0 \\
    % & \Rightarrow - \frac{2]{D_{ii}} + \rho E_{ii} D_{ii} - \rho F_{ii} D_{ii} = 0 \\
\end{align*}
leading to the following quadratic formula
$$ D_{ii}^2 - \frac{F_{ii}}{E_{ii}} D_{ii} -\frac{2}{\rho E_{ii}} = 0$$
% \begin{equation*}
%      - \frac{2]{\rho} +  E_{ii} D_{ii}^2 -  F_{ii} D_{ii} = 0
% \end{equation*}
which has the solution
$$ \widehat{D}_{ii} = \frac{\frac{F_{ii}}{E_{ii}} + \sqrt{\frac{F_{ii}^2}{E_{ii}^2} + \frac{8}{\rho E_{ii}}}}{2}$$
Then, using these diagonal elements $\hat{D}_{ii}$, it follows that
\begin{align*}
    \gamma_+  & = \argmin_{\gamma \in \mathbb{R}^{k\times k}} \Big \{ -\log (\det \gamma \gamma^T) + \frac{\rho}{2}\norm{C^{1/2} (\gamma - A)}_F^2 \Big \} \\
    & = U\widehat{D}V^T
\end{align*}
We make the final adjustment to satisfy the linear constraint. Thus, the $\gamma$ update is
\begin{equation*}
    \gamma^{(t+1)} = \gamma_+ - (\gamma_+ \textbf{1} - \textbf{a})(\textbf{1}^T C^{-1} \textbf{1})^{-1} \textbf{1}^T C^{-1}
\end{equation*}

\subsection{Proof of Proposition~\ref{thm:admm_convergence}}

\begin{proof}
    The first order conditions of the updates in Algorithm~\ref{alg:topic_vol_admm} give us
    \begin{equation}
    \begin{aligned}
        & 0 \in  \partial \norm{\cdot}_{h, \mu}(V_1^{t+1}) - \rho(\widetilde{W} \gamma^{t} - V_1^{t+1}) - \Lambda_1^t \\
        & 0 \in \mathds{1}_{G_R} (V_2^{t+1}) - \rho (\gamma^{t} - V_2^{t+1}) - \Lambda_2^t \\
        & 0 \in -2 (\gamma^{t+1})^{-T} + \rho \widetilde{W}^T ( \widetilde{W} \gamma^{t+1} - V_1^{t+1}) + \widetilde{W}^T \Lambda_1^t + \\
        & \quad \rho (\gamma^{t+1} - V_2^{t+1}) + \Lambda_2^t + \textbf{1}_K (\nu^{t+1})^T  \mt{ s.t. } \gamma^{t+1} \textbf{1} = \textbf{a}
    \end{aligned}
    \end{equation}
    Note that the first order condition for $\gamma^{t+1}$ is different as it is a equality constrained convex problem. Also, by the definitions of $\Lambda_1^{t+1}$ and $\Lambda_2^{t+1}$
    \begin{equation}
    \begin{aligned}
        & \Lambda_1^{t+1} = \Lambda_1^t + \rho(\widetilde{W} \gamma^{t+1} - V_1^{t+1})\\
        & \Lambda_2^{t+1} = \Lambda_2^t + \rho(\gamma^{t+1} - V_2^{t+1})
    \end{aligned}
    \end{equation}
    Then, combining these two sets of equations, we have that
    \begin{equation}
    \label{eq:prop1_equations}
    \begin{aligned}
        & \Lambda_1^{t+1} + \rho \widetilde{W}(\gamma^t - \gamma^{t+1}) \in \partial \norm{\cdot}_{h, \mu}(V_1^{t+1}) \\
        & \Lambda_2^{t+1} + \rho(\gamma^t - \gamma^{t+1}) \in \partial \mathds{1}_{G_R} (V_2^{t+1}) \\
        & 2 (\gamma^{t+1})^{-T} -  \textbf{1}_K (\nu^{t+1})^T= \widetilde{W}^T \Lambda_1^{t+1} + \Lambda_2^{t+1} \\
        & \frac{1}{\rho}(\Lambda_1^{t+1} - \Lambda_1^t) = \widetilde{W}\gamma^{t+1} - V_1^{t+1} \\
        & \frac{1}{\rho}(\Lambda_2^{t+1} - \Lambda_2^t) = \gamma^{t+1} - V_2^{t+1}
    \end{aligned}
    \end{equation}
    Then, let us define $(\gamma^t, V_1^t, V_2^t, \Lambda_1^t,\Lambda_2^t)_{t=1}^\infty$ be a sequence of iterates with a limit point $(\gamma^*, V_1^*, V_2^*, \Lambda_1^*,\Lambda_2^*)$. Then, by the last two equations of \eqref{eq:prop1_equations}, we have that $\widetilde{W}\gamma^* = \widetilde{W}V_2^* = V_1^*$. Therefore, the first two equations give us that
    \begin{align*}
        \Lambda_1^* & \in \partial \norm{\cdot}_{h, \mu}(V_1^*) = \partial \norm{\cdot}_{h, \mu}(\widetilde{W} \gamma^*) \\
        \Lambda_2^* & \in \partial \mathds{1}_{G_R} (V_2^*) = \partial \mathds{1}_{G_R} (\gamma^*)
    \end{align*}
    Lastly, using the third equation in \eqref{eq:prop1_equations}, it follows that
    \begin{equation*}
        \begin{aligned}
            & 2 (\gamma^*)^{-T} -  \textbf{1}_K (\nu^*)^T = \\
            & \quad \widetilde{W}^T\Lambda_1^* + \Lambda_2^* \in \partial \norm{\cdot}_{h, \mu}(\widetilde{W} \gamma^*) +  \partial \mathds{1}_{G_R} (\gamma^*)
        \end{aligned}
    \end{equation*}
    Noting that the optimality condition for $\argmin_\gamma -\log |\det(\gamma \gamma^T)| \quad \mt{s.t.} \quad \gamma \textbf{1} = \textbf{a}$ is
    $$ - 2 (\gamma^*)^{-T} +  \textbf{1}_K (\nu^*)^T = 0  \quad \mt{and $\gamma^* \textbf{1} = \textbf{a}$}$$
    We have that
    \begin{align*}
        \m{0} & = -2 (\gamma^*)^{-T} + \textbf{1}_K (\nu^*)^T + 2 (\gamma^*)^{-T} -  \textbf{1}_K (\nu^*)^T \\
        & = -2 (\gamma^*)^{-T} + \textbf{1}_K (\nu^*)^T + \widetilde{W}^T \Lambda_1^* + \Lambda_2^* \in \partial f(\gamma^*)
    \end{align*}
    and we have that $\gamma^* \textbf{1} = \textbf{a}$ by the formulation of our update for $\gamma^t$. This shows that $\gamma^*$ satisfies the optimality condition of \eqref{eq:topic_vol_obj} and thus a stationary point for $f$.
\end{proof}

\section{Simulations}
\label{appendix:sim}
We demonstrate the computational benefit as well as the accuracy of our model in terms of perplexity. The experiments are based on the simulated data from the LDA model, and we focus on the comparison to the variational EM (VEM) and Gibbs sampling to illustrate the advantages of our method. As part of the future work, we plan to compare the stochastic implementation of MVTM with GDM \citep{yurochkin_long_16} and the imporved implementations of the Gibbs sampling presented in \citet{li2014reducing} and \citet{yuan2015lightlda} at a much larger scale.

\begin{figure}[htb] \centering

\begin{subfigure}[t]{0.7\linewidth} \centering
\includegraphics[width=\linewidth]{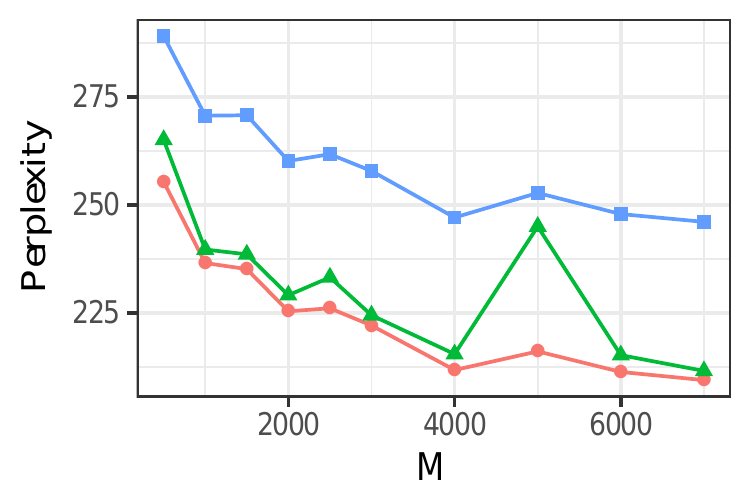}
\end{subfigure}

\begin{subfigure}[t]{0.7\linewidth} \centering
\includegraphics[width=\linewidth]{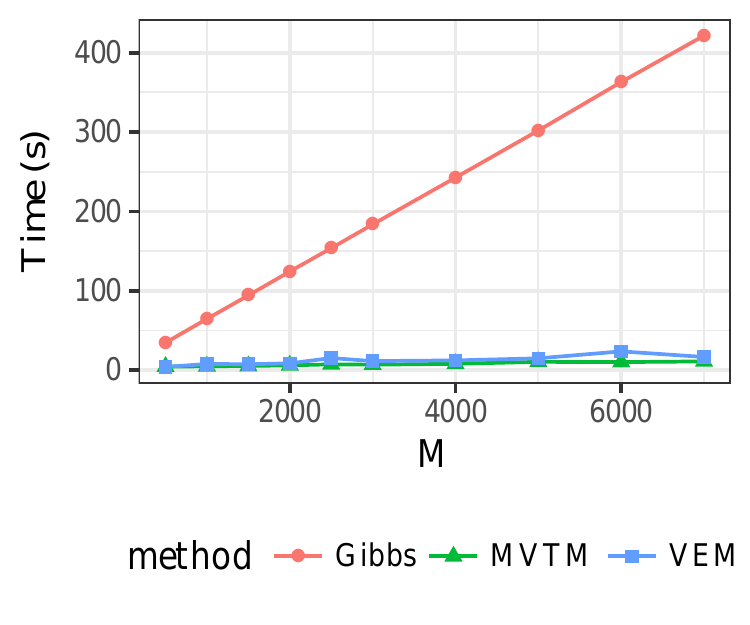}
\end{subfigure}

\caption{Perplexity of the held-out data and the corresponding time complexity of each method at varying values of the number of documents $M$ with $N_m = 1000$, $K=5$, $V=1200$, $\eta=0.1$ and $\alpha=0.1$}
\label{fig:sim_M_1000}
\end{figure}
\begin{figure}[htb] \centering

\begin{subfigure}[t]{0.7\linewidth} \centering
\includegraphics[width=\linewidth]{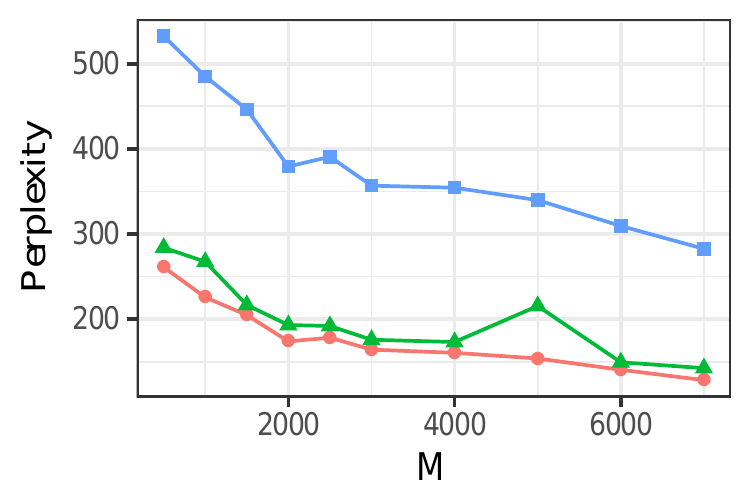}
\end{subfigure}

\begin{subfigure}[t]{0.7\linewidth} \centering
\includegraphics[width=\linewidth]{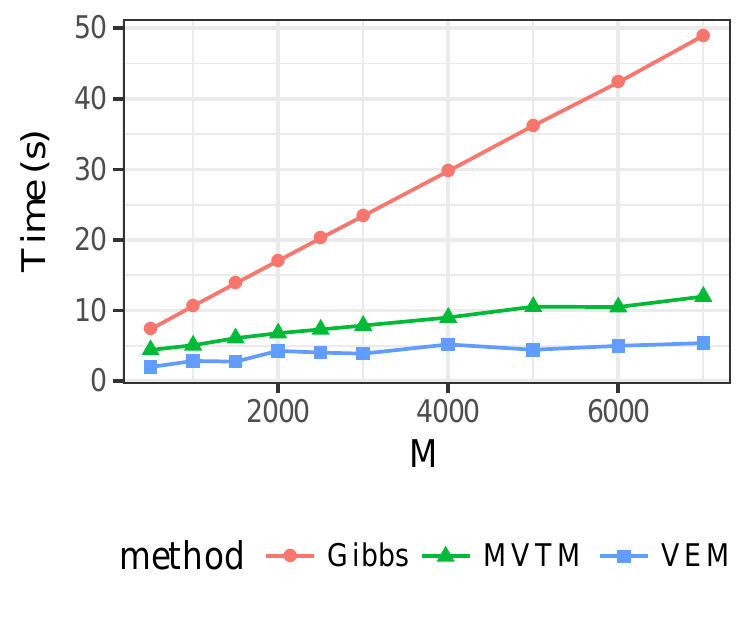}
\end{subfigure}

\caption{Perplexity of the held-out data and the corresponding time complexity of each method at varying values of the number of documents $M$ with $N_m = 100$, $K=5$, $V=1200$, $\eta=0.1$ and $\alpha=0.1$}
\label{fig:sim_M_100}
\end{figure}

We first look at the behavior of the algorithms as $M$ increases when $N_m =1000$ (Figure~\ref{fig:sim_M_100}). At $N_m = 1000$, we are working with the setting that is close to the asymptotic regime, and MVTM has the computational speed comparable to VEM and the statistical performance similar to the Gibbs sampling.

In a more challenging case with the shorter documents at $N_m = 100$, MVTM continues to perform as well as the Gibbs sampling with a little additional computational cost. This performance comparison would be of interest for the researchers who are working with shorter documents present in the modern application. As discussed in \citet{tang2014understanding} and \citet{nguyen2015posterior}, the limitation of LDA comes from the document lengths. Our results show that MVTM do not suffer from the short documents in terms of statistical performance, when the regularization parameter $\mu$ for the hinge loss is appropriately chosen. The current batch implementation, however, suffers from the number of documents present in the dataset, as it has to soft-threshold every document. This computational limitation, however, can be alleviated by the stochastic implementation as demonstrated in the stochastic implementation of the variational method in \citet{hoffman_et_al_13}.

\section{NIPS dataset Topics}
\label{appendix:nips}
\subsection{Computational Time}
\begin{figure}[h]
    \centering
    \includegraphics[width=0.9\linewidth]{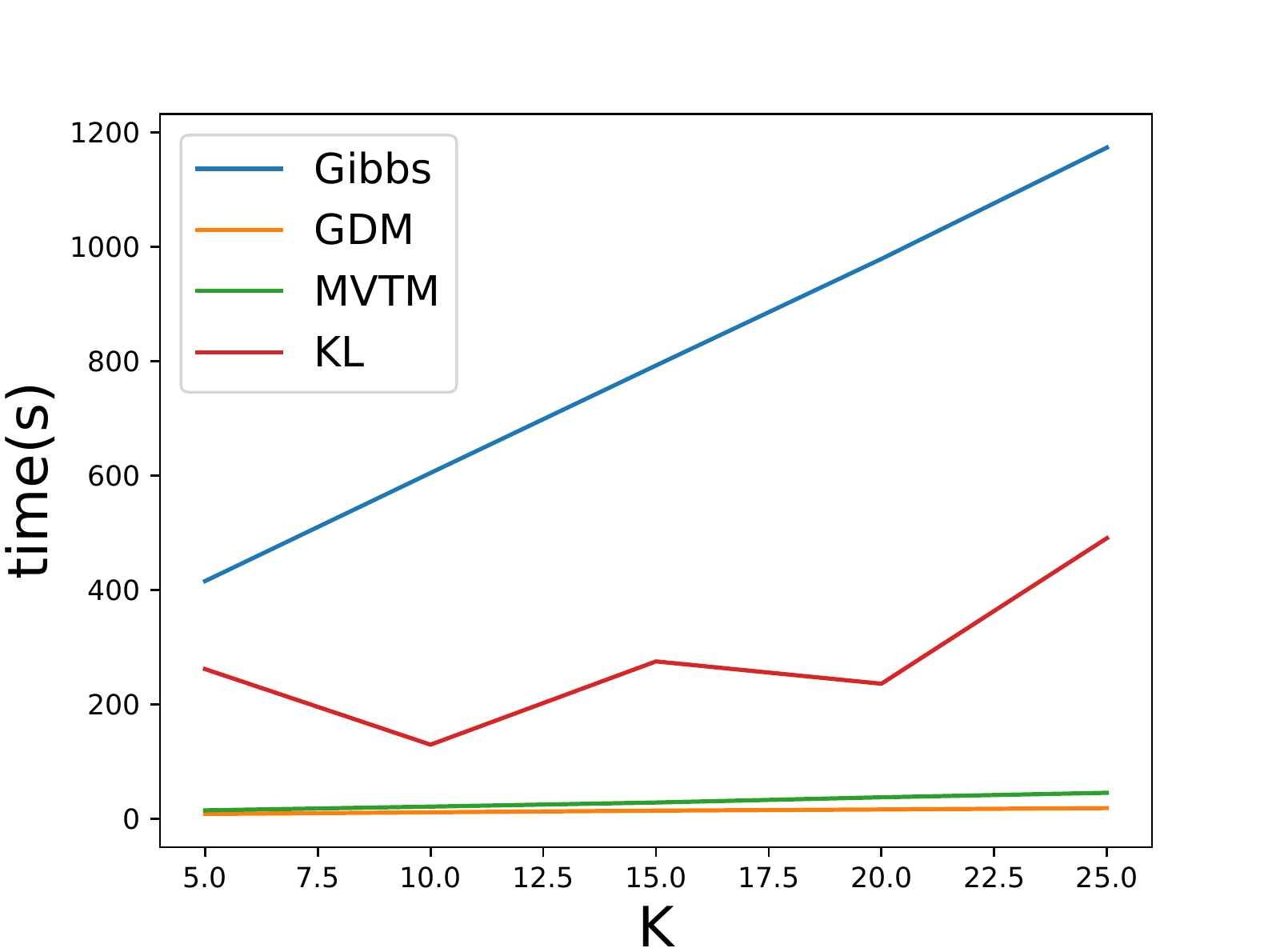}
    \caption{The computational performance of different algorithms as a function of the number of topics. NIPS dataset includes 1491 documents and 4492 unique words.}
    \label{fig:nips_time_complexity}
\end{figure}

Figure~\ref{fig:nips_time_complexity} shows the time complexities of different algorithms on the NIPS dataset as we increase the number of topics. Compared to GDM, the proposed MVTM improvement on performance comes at a little computational cost. RecoverKL could achieve similar computational speed if the anchor words are provided. However, when we include the computational cost of finding the anchor words, GDM and MVTM show computational advantages over RecoverKL.

\vfill
\pagebreak
\subsection{Top 10 topics}
\begin{sidewaystable}
\begin{tabular}{llllllllll}
\hline
 Topic 1  & Topic 2 & Topic 3     & Topic 4    & Topic 5      & Topic 6  & Topic 7      & Topic 8  & Topic 9       & Topic 10      \\
 \hline
 neuron   & input   & training    & training   & algorithm    & unit     & model        & network  & function      & learning      \\
 network  & output  & set         & error      & learning     & network  & data         & neural   & set           & system        \\
 input    & system  & network     & set        & data         & input    & parameter    & system   & approximation & control       \\
 model    & circuit & recognition & data       & problem      & hidden   & distribution & problem  & result        & function      \\
 pattern  & signal  & data        & cell       & weight       & weight   & system       & training & linear        & action        \\
 neural   & neural  & algorithm   & input      & method       & output   & object       & control  & bound         & algorithm     \\
 synaptic & network & vector      & network    & function     & layer    & gaussian     & dynamic  & number        & task          \\
 learning & chip    & learning    & classifier & distribution & learning & likelihood   & unit     & point         & reinforcement \\
 cell     & weight  & classifier  & weight     & vector       & pattern  & cell         & result   & network       & error         \\
 spike    & analog  & word        & test       & parameter    & training & mixture      & point    & threshold     & model         \\
\hline
\end{tabular}
\caption{Top 10 MVTM topic for NIPS dataset}
\end{sidewaystable}

\begin{sidewaystable}
\begin{tabular}{llllllllll}
\hline
 Topic 1  & Topic 2 & Topic 3     & Topic 4        & Topic 5        & Topic 6  & Topic 7   & Topic 8   & Topic 9      & Topic 10      \\
 \hline
 neuron   & circuit & recognition & set            & model          & network  & function  & image     & model        & learning      \\
 cell     & signal  & speech      & training       & memory         & input    & algorithm & object    & data         & control       \\
 model    & system  & word        & data           & representation & unit     & learning  & images    & distribution & system        \\
 input    & neural  & system      & algorithm      & node           & weight   & point     & field     & gaussian     & action        \\
 activity & analog  & training    & error          & rules          & neural   & vector    & map       & parameter    & model         \\
 synaptic & chip    & hmm         & performance    & tree           & output   & result    & visual    & mean         & dynamic       \\
 pattern  & output  & character   & classifier     & structure      & learning & case      & motion    & algorithm    & policy        \\
 response & current & model       & classification & level          & training & problem   & feature   & probability  & algorithm     \\
 firing   & input   & network     & number         & graph          & layer    & parameter & direction & method       & reinforcement \\
 cortex   & neuron  & context     & learning       & rule           & hidden   & equation  & features  & component    & problem       \\
\hline
\end{tabular}
\caption{Top 10 Gibbs topic for NIPS dataset}
\end{sidewaystable}

\begin{sidewaystable}
\begin{tabular}{llllllllll}
\hline
 Topic 1  & Topic 2 & Topic 3     & Topic 4    & Topic 5        & Topic 6  & Topic 7      & Topic 8     & Topic 9     & Topic 10      \\
 \hline
 neuron   & input   & word        & data       & image          & network  & model        & cell        & learning    & learning      \\
 network  & output  & speech      & set        & images         & unit     & data         & visual      & algorithm   & control       \\
 spike    & weight  & recognition & training   & object         & neural   & parameter    & motion      & function    & model         \\
 synaptic & neural  & system      & error      & point          & weight   & likelihood   & direction   & problem     & system        \\
 input    & network & training    & function   & features       & hidden   & mixture      & response    & action      & task          \\
 pattern  & net     & character   & vector     & graph          & training & distribution & orientation & policy      & movement      \\
 firing   & chip    & hmm         & method     & representation & output   & algorithm    & neuron      & optimal     & controller    \\
 model    & layer   & speaker     & classifier & feature        & input    & set          & model       & gradient    & motor         \\
 activity & analog  & context     & kernel     & information    & error    & gaussian     & frequency   & convergence & dynamic       \\
 neural   & bit     & network     & gaussian   & recognition    & function & variables    & field       & step        & reinforcement \\
\hline
\end{tabular}
\caption{Top 10 GDM topic for NIPS dataset}
\end{sidewaystable}

\end{document}